\documentclass{article}
\usepackage{arxiv}
\usepackage[utf8]{inputenc} 
\usepackage[T1]{fontenc}    
\usepackage{hyperref}       
\usepackage{url}           
\usepackage{booktabs}       
\usepackage{amsfonts}      
\usepackage{nicefrac}       
\usepackage{microtype}     
\usepackage{lipsum}		
\usepackage{graphicx}
\usepackage{doi}
\usepackage{todonotes}
\usepackage{wrapfig}
\usepackage{titlesec}
\usepackage{paralist}

\titlespacing{\section}{0pt}{5ex}{2ex}
\titlespacing{\subsection}{0pt}{3ex}{1ex}
\titlespacing{\subsubsection}{0pt}{2ex}{2ex}

\renewcommand{\undertitle}{}

\begin{document}

\title{Determinants of LLM-assisted Decision-Making}

\author{ 
\href{https://orcid.org/0009-0002-2014-4274}{\includegraphics[scale=0.06]{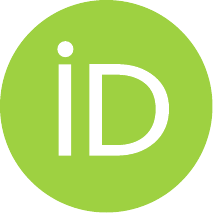}\hspace{1mm}Eva Eigner} and 
\href{https://orcid.org/0000-0002-0589-204X}{\includegraphics[scale=0.06]{orcid.pdf}\hspace{1mm}Thorsten Händler} \\ 
	Ferdinand Porsche Mobile University of Applied Sciences (FERNFH)\\
	Wiener Neustadt, Austria \\
	\texttt{eva.eigner@fernfh.ac.at; thorsten.haendler@fernfh.ac.at} \\
 }

\date{}

\hypersetup{
pdftitle={Determinants of LLM-assisted Decision-Making},
pdfsubject={cs.AI, cs.HC},
pdfauthor={Eva Eigner, Thorsten Händler},
pdfkeywords={Decision-making, human-computer interaction, large language models (LLMs), conversational AI, psychological determinants, dependency framework, decision-support systems, prompt enhgineering, over-reliance},
}

\maketitle

\vspace{-2mm}

\begin{abstract}
Decision-making is a fundamental capability in everyday life. Large Language Models (LLMs) provide multifaceted support in enhancing human decision-making processes. However, understanding the influencing factors of LLM-assisted decision-making is crucial for enabling individuals to utilize LLM-provided advantages and minimize associated risks in order to make more informed and better decisions. This study presents the results of a comprehensive literature analysis, providing a structural overview and detailed analysis of determinants impacting decision-making with LLM support. In particular, we explore the effects of technological aspects of LLMs, including transparency and prompt engineering, psychological factors such as emotions and decision-making styles, as well as decision-specific determinants such as task difficulty and accountability. In addition, the impact of the determinants on the decision-making process is illustrated via multiple application scenarios. Drawing from our analysis, we develop a dependency framework that systematizes possible interactions in terms of reciprocal interdependencies between these determinants. Our research reveals that, due to the multifaceted interactions with various determinants, factors such as  trust in or reliance on LLMs, the user's mental model, and the characteristics of information processing are identified as significant aspects influencing LLM-assisted decision-making processes. Our findings can be seen as crucial for improving decision quality in human-AI collaboration, empowering both users and organizations, and designing more effective LLM interfaces. Additionally, our work provides a foundation for future empirical investigations on the determinants of decision-making assisted by LLMs.
\end{abstract}

\vspace{-2mm}

\section{Introduction}
Every day, individuals are confronted with a variety of situations that require decision-making. Consequently, the ability to make decisions by reflecting relevant information and weighing up available decision options in an efficient way is a critical and fundamental capability \cite{Mather2006ARO}. For many important decisions, both personal and professional, individuals seek the advice of experts. While in the past advice commonly has been sought from human experts, today advice based on artificial intelligence (AI) is increasingly emerging \cite{chong2022human,Schemmer2022ShouldIF,lu2024does}.
AI-assisted decision-making is supposed to lead to quicker and improved decision outcomes \cite{Angerschmid2022Fairness}. Hence, its use in decision-making can be seen as one of the most significant applications of AI \cite{DuanArtificialIntelligence2019}.

Large Language Models (LLMs) offer versatile assistance in decision-making processes. For instance, their ability to process and summarize extensive text data \cite{Liu2023TrustworthyLA} enables decision-makers to comprehend key insights swiftly. Moreover, LLMs are adept at idea generation \cite{Girotra2023Ideas} and are capable of generating different solutions \cite{yang2023large}, enhancing the creation of various alternatives in decision-making. They can also identify patterns \cite{jin2023time} and analyze historical data \cite{rane2023chatgpt}, thus potentially providing support in the analysis of decision situations and evaluation of alternatives. Additionally, LLMs demonstrate the capability to adopt personas of various characters and engage in social interactions with each other \cite{park2022social} and can simulate debates featuring different opinions \cite{taubenfeld2024systematic}. Through the ability to incorporate diverse perspectives and simulate discussions, LLMs empower decision-makers to systematically explore a multitude of scenarios and potential outcomes. Moreover, LLMs exhibit a high degree of rationality in decision-making tasks \cite{chen2023emergence}, implying that LLMs hold the potential to enhance human decision-making processes by providing reasoned outputs. Thus, in the realm of \textit{AI-assisted decision-making} \cite{Tejeda2022AI-assisted, Steyvers2023Three}, LLMs can be seen as powerful and promising tools due to their multifaceted capabilities.

Nevertheless, the increased capabilities of  LLMs are associated with heightened risks \cite{ji2023ai}. Undesirable behaviors exhibited by LLMs encompass, for instance, generating nonfactual or untruthful information (\textit{hallucinations}) \cite{bang2023multitask}, reiterating a user's presented viewpoints (\textit{sycophancy}) \cite{perez2022discovering}, providing false rationalizations that diverge from the true reasons behind the LLMS' outputs (\textit{unfaithful reasoning}) \cite{turpin2023language}, and employing deception because LLMs have rationalized that it can advance a particular objective (\textit{strategic deception}) \cite{park2023ai}. A persistent concern is the potential loss of human control over AI systems, permitting these systems to pursue objectives that may contradict individuals' interests \cite{park2023ai}. This risk is exacerbated by the autonomous capabilities present in current LLMs \cite{liu2023agentbench, kinniment2023evaluating}. Furthermore, LLMs may unintentionally process harmful information inherent to their training data, including biases, discrimination, and toxic content \cite{Weidinger2021EthicalAS}. Thus, LLMs may generate content or engage in behaviors that humans may wish to avoid due to their undesirability or potential risks \cite{shen2023large}. 
Hence, various risks are inherent to LLMs deciding and acting autonomously, particularly concerning the extent to which AI systems are aligned with human values, intentions and goals \cite{russell2019human}.

One possible solution to mitigate these risks is to form hybrid human-AI teams \cite{Bosch2019Six}, in which the AI supports the individual in the decision-making process by making recommendations or suggestions, but the individual remains responsible for the final decision \cite{Bansal2020DoesTW}. A decision-making process that engages humans and AI can profit from the strengths of each party \cite{Holstein2021DesigningFH}. Humans can effectively monitor unpredictable or undesirable behavior exhibited by AI models and, crucially, integrate vital contextual information. The integration of AI into decision-making processes allows the processing of more complex patterns and larger amounts of data than humans can handle \cite{Kelly2023CapturingHM}. The synergy between humans and AI, often referred to as \textit{complementarity}, aims to achieve performance superior to that of either humans or AI in isolation \cite{Bansal2020DoesTW}. To enhance complementarity and increase the efficacy and efficiency of LLM-assisted decision-making, it is crucial to understand the underlying \textit{determinants}. 

Determinants are generally referred to as causal factors, and variations in these factors lead to systematic changes in behavior \cite{Bauman2002TowardAB}. Behavioral determinants relate to any condition influencing human behavior and the interaction of such conditions \cite{def2024determinant}, providing explanations and predictions for human decision-making and behavior \cite{Vrabel2017Conscious}. In the context of LLM-assisted decision-making, we refine these definitions to categorize determinants as causal factors and conditions that influence and predict human decision-making behavior with the assistance of LLMs. Our definition further encompasses the interaction among these factors, indicating that they influence each other. Recognizing the determining factors at play and being aware of their influence enables individuals to use LLMs' capabilities more effectively in decision-making. These determinants may span psychological, technological, and decision-specific aspects. Comprehending these factors, along with their interactions, is significant for optimizing the synergy between human expertise and abilities of LLMs, thereby enhancing the efficiency and effectiveness of decision-making processes.

However, research on determinants of LLM-assisted decision-making and their interactions is scarce. To the best of our knowledge, there is no comprehensive overview of the determinants of either LLM-assisted or AI-assisted decision-making. Previous studies have primarily focused on singular selected factors influencing LLM- or AI-assisted decision-making. For instance, Liao and Vaughan \cite{Liao2023AITI} highlight the importance of transparency in LLM-assisted decision-making. In the realm of AI-assisted decision-making, the impact of explanations on over-reliance on AI during the decision-making process has already been investigated \cite{Buccinca2021ToTO}. However, so far there is no comprehensive analysis investigating the characteristics and interplay of the various factors that influence AI-assisted decision making.


This paper aims to bridge this gap by developing a comprehensive understanding of the determinants that specifically influence LLM-assisted decision-making by providing the following \textbf{contributions}:

\begin{compactitem}
    \item [1.]We present a \textit{structural overview} and detailed analysis of technological, psychological, and decision-specific factors determining LLM-assisted decision-making as result of an integrative literature review. 
    \item [2.]Drawing from this analysis, we develop a \textit{dependency framework} systematizing the potential interactions and interdependencies between these determinants. 
    \item [3.]Furthermore, we demonstrate the utility of our work by illustrating its application in the context of multiple exemplary scenarios.
\end{compactitem}

Hence, this paper significantly contributes to advancing the comprehension of factors influencing human decision-making with the support of LLMs and their interactions. Awareness of these determinants and their interdependencies can empower decision-makers and organizations to improve the quality of LLM-assisted decision-making, thereby mitigating potential risks such as over-reliance on LLMs, which occurs when humans fail to correct erroneous AI recommendations \cite{Vasconcelos2022ExplanationsCR}. Understanding the determinants that impact LLM-assisted decision-making can promote user empowerment by leveraging the advantages. When users are cognizant of these factors, they can formulate more precise queries to LLMs, obtaining more relevant and accurate information, thus enhancing decision quality. Moreover, this knowledge enables users to critically assess LLM-generated output, potentially leading to improved decision-making. For instance, being aware of how psychological factors interact with the technological aspects of LLMs allows users to comprehend how their expectations, experiences, trust in AI, and biases collectively shape decisions. This awareness facilitates more thoughtful and informed decision-making processes.

The structural overview of determinants of LLM-assisted decision-making and the dependency framework of the interactions among these factors can serve as basis for supporting personnel development within organizations. When organizations comprehend these determinants, they are better equipped to enhance their personnel development strategies. By incorporating this framework into training initiatives, organizations can design targeted training programs for their employees. Understanding the factors influencing LLM-assisted decision-making is also critical for user design. Knowledge of these determinants enables designers to create more tailored design interfaces and interactions that resonate with users, increasing engagement and satisfaction. Moreover, awareness of influencing factors helps in anticipating user needs and potential challenges.

\textbf{Structure.} The remainder of this paper is structured as follows: In Section \ref{sec:background}, we discuss the background and related work on decision-making and LLMs. Section \ref{sec:approach-overview} provides an overview of the approach, outlining the applied methodology. In Section \ref{sec:tech-determinants}, technological determinants of LLM-assisted decision-making are analyzed, while Section \ref{sec:psych-determinants} details psychological determinants. Section \ref{sec:ds-determinants} extends these determinants by discussing decision-specific factors. In Section \ref{sec:framework}, the resulting dependency framework, derived from the analysis in the previous sections, is presented. Section \ref{sec:discussion} discusses implications for decision-makers and organizations, limitations and further potentials of the applied approach. Finally, Section \ref{sec:conclusion} concludes the paper.

\section{Background}
\label{sec:background}
In this section, we give an overview of relevant basics from the fields of human decision-making (see Section \ref{subsec:human decision-making}), decision-support systems (see Section \ref{subsec: AI-assisted Decision Making}), large language models (LLMs) (see Section \ref{subsec: LLM}) as well as the decision-making process assisted by LLMs (see Section \ref{subsec: LLM-assisted DMP}).

\subsection{Human Decision-Making}
\label{subsec:human decision-making}
Human decision-making can be understood as the conscious process of evaluating different options and choosing the most adequate one to achieve one or more defined goals, relying on the individual's skills, values, preferences, and beliefs \cite{MorelliDecisionMaking2022} It is further influenced by situational and contextual variables, including factors like time pressure \cite{Shepherd2013TheInfluence}. Furthermore, a decision is defined as “goal-directed behavior made by the individual in response to a certain need, with the intention of satisfying the motive that the need occasions” \cite[p.~86]{Jabes1978individual}. 

Simon \cite{simon1977new} proposed a decision process consisting of the following three phases:
\begin{compactitem}
\item [1.]\textit{Intelligence}: In this initial stage, the decision maker recognizes the problem and the need to make a decision and gathers information about the problem situation. 
\item [2.]\textit{Design}: The design phase involves systematically structuring the problem, establishing specific criteria, and identifying a range of alternatives aimed at resolving the issue at hand.
\item [3.]\textit{Choice}: In the choice phase, the decision-maker selects the most optimal alternative that aligns with the defined criteria and subsequently makes the final decision.
\end{compactitem}
The decision-making process is a complex and continuous task, often being partly iterative, where phases can overlap, and the decision-maker might revisit previous stages \cite{simon1997administrative}.
In Section \ref{subsec: LLM-assisted DMP}, we detail how LLMs can assist during various decision-making phases.

\subsection{Decision Support Systems and AI-assisted Decision-Making}
\label{subsec: AI-assisted Decision Making}

The primary goal of an effective Decision Support System (DSS) is “to guide and direct the decision-maker towards a better solution” \cite[p. 356]{Todd1999EvaluatingTI}. DSSs are designed to perform various functions, including managing the overflow of information and knowledge, as well as assisting decision makers in clarifying their judgments and preferences \cite{razmak2015decision}. DSSs have the capability to assist in overcoming human cognitive limitations by integrating diverse information sources, providing intelligent access to relevant knowledge, and facilitating the decision-making process \cite{Mir1979Decision}. When AI methods are employed to create options, the resulting system is known as an Intelligent Decision Support System (IDSS) \cite{Phillips-Wren2013Intelligent}. IDSSs are tools designed to assist in decision-making processes characterized by uncertainty or incomplete information, or when decisions include risks \cite{Jantan2010Intelligent}. The hope is that such tools will lead to an enhanced efficiency of human decision-making processes \cite{fogliato2022goes}. 

Human-AI decision-making, also referred to as AI-assisted decision-making, involves scenarios where an AI model supports the user in making a final judgment or decision, frequently viewed as a kind of collaboration between humans and AI systems \cite{Chen2023UnderstandingTR}. Ultimately, the human decision-maker makes the final decision \cite{Steyvers2023Three}. In AI-assisted human decision-making, the following cycle is typically repeated: (1) receiving input from the environment, (2) the AI suggesting a (possibly erroneous) action, (3) the human making a decision based on the AI's input, and (4) the environment providing feedback, which the person learns when to trust the AI's recommendation \cite{Bansal2019BeyondAT}.

\subsection{Large Language Models}
\label{subsec: LLM}
Large Language Models (LLMs), such as \textit{GPT-4}, are subsumed under the category of generative AI \cite{Chang2023ASO, Kasneci2023ChatGpt, GozaloBrizuela2023ChatGPTIN}. These advanced Transformer-based language models with hundreds of billions or more parameters are trained on extensive data \cite{Shanahan2022TalkingAL}. Studies have indicated that scaling significantly enhances the model capacity of LLMs \cite{Brown2020LanguageMA, Chowdhery2022PaLMSL}. LLMs demonstrate remarkable abilities in understanding natural language and solving complex text generation tasks \cite{zhao2023survey}. Engineered to comprehend and produce natural language, LLMs are capable of performing a wide array of natural language tasks, including automatic summarizing, machine translation, and question answering \cite{Shen2023ChatGPT}.

In contrast to conventional (smaller) language models, LLMs are characterized by so-called \textit{emergent abilities}. An emergent ability is defined as an ability that "is not present in smaller models but is present in larger models” and is related to specific complex tasks \cite{Wei2022EmergentAO}. An example of an emergent ability is step-by-step reasoning. For small language models, solving complex tasks involving multiple reasoning steps, such as mathematical word problems, has posed difficulties. In contrast, LLMs can tackle such tasks, for instance by using the \textit{Chain-of-Thought (CoT)} prompting strategy \cite{Wei2022ChainOT}. Furthermore, through \textit{Instruction Tuning}, LLMs are capable to follow task instructions for new tasks without relying on explicit examples, enhancing their overall ability for generalization \cite{zhao2023survey}. Additionally, LLMs are able to execute unfamiliar tasks solely by reading task instructions without requiring a few-shot examples, a capability referred to as \textit{Instruction Following} \cite{Wei2022EmergentAO}.

However, not only emergent abilities can arise, but also risks. For example, LLMs can perpetuate stereotypes and social biases, leading to unfair discrimination. In addition to this, LLMs might provide false or misleading information that can be harmful, especially in critical areas like legal or medical advice. Another risk lies in presenting LLMs as "human-like" which potentially leads users to overestimate their capabilities \cite{weidinger2022taxonomy}. A comprehensive overview of challenges, limitations and risks of LLMs is provided in Section \ref{subsec: Challenges}.

\subsection{LLM-assisted Decision-Making Process}
\label{subsec: LLM-assisted DMP}
\begin{figure*}[htb]
	\centering
	\includegraphics[width=1.0\textwidth]{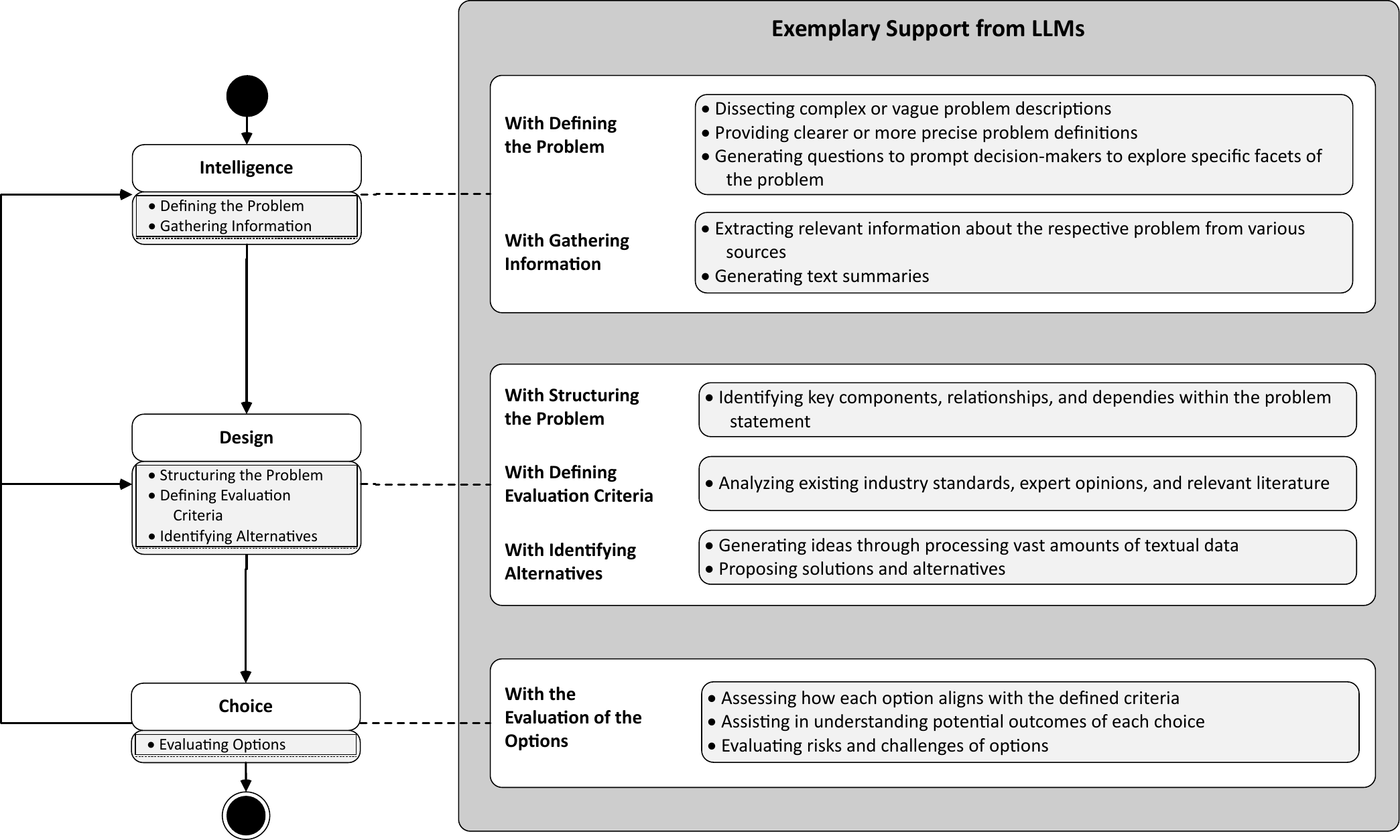}
	\caption{Key stages in the decision-making process oriented to Simon \cite{simon1977new} extended by LLM support options.}
	\label{fig:dm-process}
\end{figure*}

As mentioned in Section \ref{subsec:human decision-making}, the decision-making process can be structured into three main phases: intelligence, design, and choice \cite{simon1977new}. As depicted in Figure\ \ref{fig:dm-process}, LLMs can provide assistance in each stage of this process. 

In the \texttt{Intelligence} Phase of decision-making, LLMs can assist in both defining the problem and gathering essential information about the situation. Concerning problem definition, LLMs are capable of dissecting complex or vague problem descriptions, providing clearer and more precise definitions. Moreover, these models can generate questions that might guide decision-makers, prompting them to explore specific facets of the issue at hand. In the realm of information gathering, LLMs can assist by accessing a variety of sources and gathering relevant information about the respective problem. Furthermore, they can generate text summaries and extract possible key details, thereby enhancing the decision-maker's understanding of the core information.

In the \texttt{Design} Phase of decision-making, LLMs can provide assistance in structuring the problem by identifying key components, relationships, and dependencies within the problem statement. Moreover, LLMs may support the decision-maker in defining specific and relevant criteria for evaluating potential alternatives, for instance, by analyzing existing industry standards, expert opinions, and relevant literature. LLMs can contribute to identifying alternatives aimed at resolving the problem by generating ideas through processing vast amounts of textual data. Hence, they are capable of proposing solutions and alternatives that decision-makers might not have considered otherwise.

In the \texttt{Choice} Phase of decision making, LLMs can support the decision-maker by evaluating and comparing different options based on defined criteria. They are capable to process vast amounts of textual data to assess how each option aligns with the defined criteria, enabling decision-makers to make data-driven choices. Based on the provided data, LLMs can simulate different scenarios by including various choices. This can help decision-makers understand the potential outcomes of each choice and enables them to select options with favorable consequences. As LLMs analyze historical data to assess potential risks associated with each option, LLMs can aid decision-makers in making informed decisions that consider potential challenges.

\vspace{-2mm}

\section{Methodological Approach}
\label{sec:approach-overview}
\begin{figure}
    \includegraphics[width=1.0\textwidth]{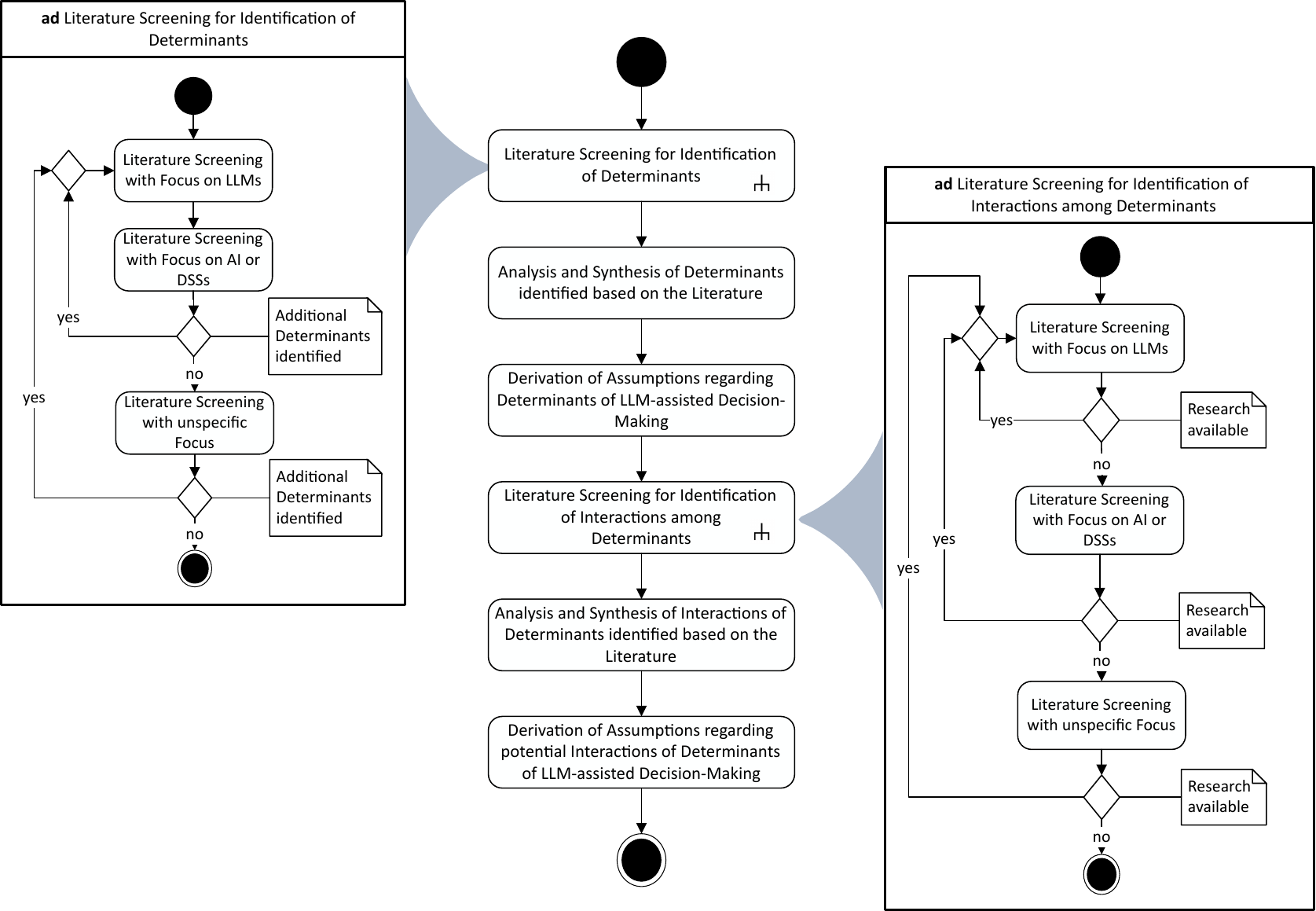}
    \caption{Process of the applied methodological approach.}
    \label{fig: methodology}
\end{figure}

\vspace{-1mm}
Considering the nascent nature of LLM-assisted decision-making, an \textit{integrative} approach was employed as the method for the literature review, as this type of review is designed to address new and emerging topics \cite{Snyder2019Literature}. Selecting this methodology was guided by the objective of this work, i.e., the development of a framework for determinants of LLM-assisted decision-making and their interactions (\textit{dependency framework}). This aligns with the potential contribution of the integrative review, whose purpose is to evaluate and synthesize literature, generating advancements in knowledge and new theoretical frameworks \cite{torraco2005writing}. Adhering to the criteria of an integrative review, our sample comprised research articles and books \cite{Snyder2019Literature}, with the literature search being conducted in an interdisciplinary manner due to the nature of the topic. Consistent with the principles of an integrative review, the applied research process, especially the literature selection \cite{Snyder2019Literature}, is documented below. To identify determinants influencing LLM-assisted decision-making and their interactions, the process was structured into several key stages. 
The process is illustrated in Figure\ \ref{fig: methodology} as an activity diagram of the \textit{Unified Modeling Language (UML2)} \cite{UML2017}.

\begin{enumerate}
    \item \textbf{Literature Screening for Identification of Determinants:} 
    The primary objective of this stage was to conduct a thorough literature review to identify determinants potentially significant in influencing LLM-assisted decision-making. This literature screening represents a sub-process consisting of multiple steps (see left-hand side in Fig.\ \ref{fig: methodology}). In particular, in step 1, this process involved gathering theories and research specifically related to factors influencing decision-making facilitated by LLMs. In step 2, the literature search was expanded to include determinants within the context of decision-making with the assistance of AI or DSSs. The identification of additional potential determinants prompted a literature review to assess the presence of research findings related to these factors within the context of LLM-assisted decision-making. In step 3, these determinants were complemented by factors widely acknowledged in the literature as important for the decision-making process in general or considered relevant by the authors. If further potential determinants were discerned (in addition to steps 1 and 2), a literature screening was conducted to appraise the existence of research concerning these factors within the realm of decision-making with assistance of LLMs.
    
    \item \textbf {Analysis and Synthesis of Determinants identified based on the Literature:} The aim of this stage was to acquire a comprehensive understanding of potential determinants of LLM-assisted decision-making. To achieve this, an analysis and synthesis of the research on the previously identified determinants of decision-making with or without the support of LLMs, AI and DSSs were conducted in order to to discern patterns, consistencies, and divergences within the gathered information. 

    \item \textbf {Derivation of Assumptions regarding potential Determinants of LLM-assisted Decision-Making:} The desired outcome of this stage was to obtain a comprehensive understanding of determinants of LLM-assisted decision-making. Based on the previously conducted analysis and synthesis, assumptions were derived concerning potential influencing factors on LLM-assisted decision-making. Furthermore, conclusions were drawn regarding the specific impact of the identified factors on decision-making with the support of LLMs.
    
    \item \textbf{Literature Screening for Identification of Interactions among Determinants:} In this stage, a literature search was conducted aiming to explore interactions between the previously identified determinants of LLM-assisted decision-making. This screening again represents a sub-process consisting of multiple steps (see right-hand side in Fig.\ \ref{fig: methodology}). 
    For each influencing factor, a literature screening was carried out to identify research on interactions with each of the other determinants. Initially, the focus was on determining whether interactions could be found in the literature within the context of LLM-assisted decision-making. If not, the search was extended to decision-making with the support of AI and DSSs. In the absence of domain-specific literature, a broader search was conducted to include general literature on the interactions of these determinants. 
    
    \item \textbf{Analysis and Synthesis of Interactions of Determinants identified based on the Literature:} 
    In this stage, the goal was to obtain a thorough overview of the previously identified interactions of determinants influencing decision-making processes, whether with or without the assistance of LLMs, AI, or DSSs. Consequently, the insights acquired from the literature review were analyzed and synthesized, with a focus on identifying similarities and contradictions within literature. 
    
   \item \textbf{Derivation of Assumptions regarding potential Interactions of Determinants of LLM-assisted Decision-Making:} The goal of this stage was to illustrate potential interactions among the identified determinants of LLM-assisted decision-making. Building on the analysis and synthesis of the preceding phase, assumptions were derived to explain how interactions among the identified determinants might manifest in the context of LLM-assisted decision-making.
\end{enumerate} 


In Section \ref{subsec: Determinants Structure}, we outline the identified technological, psychological, and decision-specific determinants that our study focuses on. Additionally, we introduce means to structure the determinants and their interactions, which also includes notations used for figures and symbols. In Section \ref{subsec: Application Scenarios}, we present various scenarios to illustrate the determinants' potential impact in LLM-assisted decision-making.

\vspace{3mm}

\subsection{Determinants Structure}
\begin{figure*}[htb]
	\centering
	\includegraphics[width=1.0\textwidth]{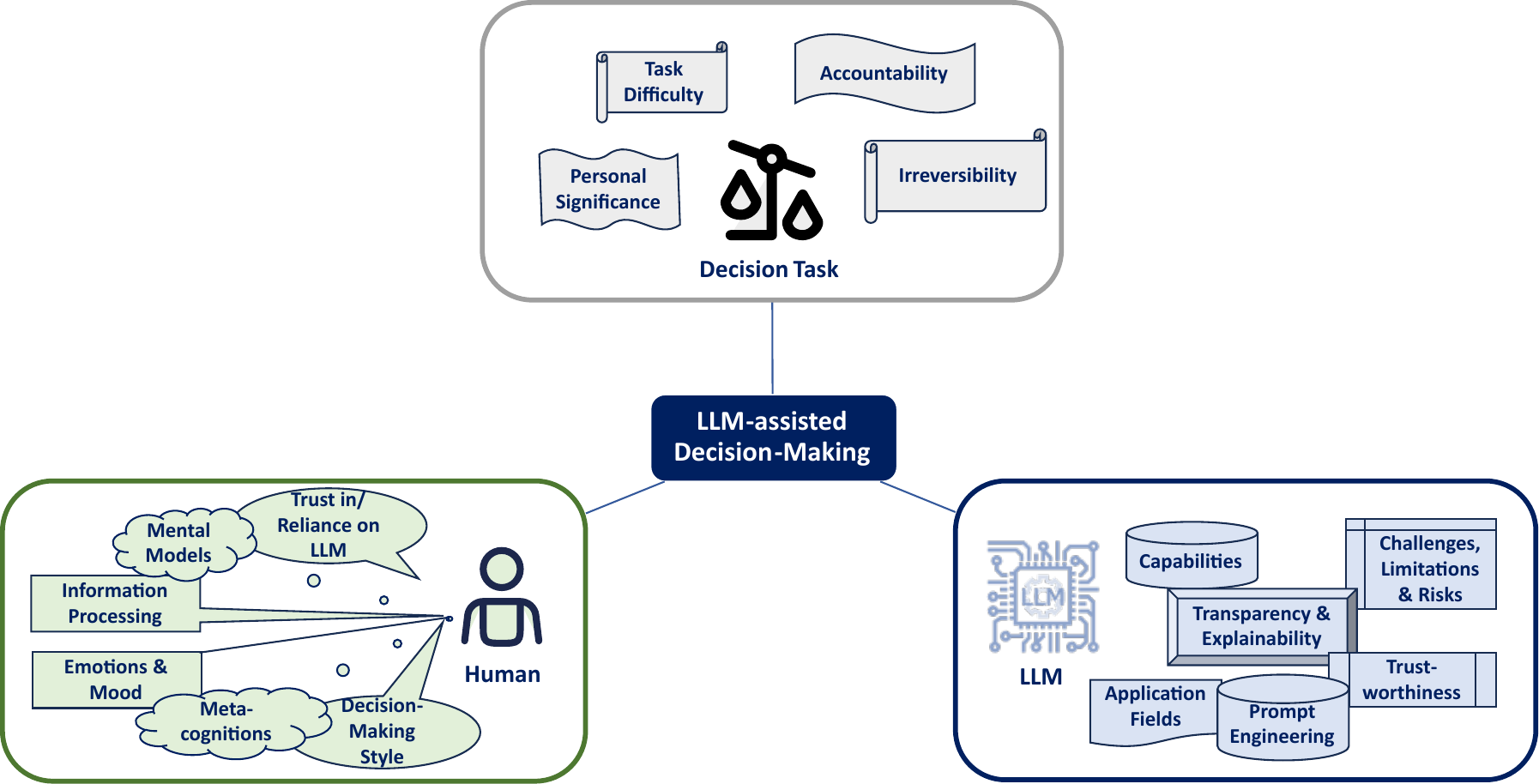}
	\caption{Schematic overview of addressed determinants of LLM-assisted decision-making.}
	\label{fig: overview determinants}
\end{figure*}

As illustrated in Figure\ \ref{fig: overview determinants}, we focus on the interactions between psychological (\texttt{Human} 
related), technological (\texttt{LLM} related), and decision-specific determinants (\texttt{Decision Task} related). 

\label{subsec: Determinants Structure}

The inclusion of technological factors is driven by their direct relevance to the technological capabilities and limitations inherent to LLMs. For example, understanding the abilities of LLMs is crucial for evaluating their applicability and effectiveness in decision-making. Psychological factors play a central role in comprehending human-technology interaction. Therefore, the adoption and effectiveness of LLM-assisted decision-making are significantly influenced by human factors, including trust in technology, cognitive biases, and decision-making styles. By examining psychological aspects, we aim to gain insights into how users interact with, interpret, and are influenced by the outputs of LLMs. This understanding is vital for optimizing their role in decision support. Decision-specific factors address the characteristics of a decision task, such as its complexity. By focusing on these factors, this paper aims to provide a comprehensive understanding of how LLMs can be adapted and applied effectively in various decision-making contexts, from routine operational decisions to complex strategic planning.

Excluding environmental factors, such as the influence of social norms and regulations, from our analysis is based on a key rationale: the relative difficulty individuals or organizations to exert significant influence over these factors and that such determinants are beyond the immediate control or influence of individuals or organizations. Unlike technological, decision-specific or psychological factors, which can be more easily changed or adapted through specific interventions or strategies, environmental factors are more resistant to change and not easily altered by the actions of a single entity or group.

For the structuring the determinants and their sub-determinants of LLM-assisted decision-making as well as for visualizing the interdependencies between the determinants, feature diagrams are utilized. Predominantly employed in software engineering, a feature diagram visually represents a feature model \cite{Kang1990Feature}, describing the hierarchical structure of system features and the relationships between a parent feature and its sub-features \cite{Batory2005Feature}. In addition, further interdependiencies between features can be represented. 
For the purpose of our analysis, the following notations of feature diagrams are employed to structure the interactions and interdependencies between determinants; also see \cite{Kang1990Feature}:
\\
\\
\vspace{0,1em}
\begin{minipage}{0.25\textwidth}
  \includegraphics[width=0.2\linewidth]{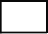}
\end{minipage}
\hspace{-4.5em}
\begin{minipage}{0,9\textwidth}
Determinant.
\end{minipage}

\vspace{0,1em}
\begin{minipage}{0.25\textwidth}
  \includegraphics[width=0.25\linewidth]{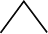}
\end{minipage}
\hspace{-4.5em}
\begin{minipage}{0,9\textwidth}
All sub-determinants have an impact on LLM-assisted decision-making.
\end{minipage}

\vspace{0,1em}
\begin{minipage}{0.25\textwidth}
  \includegraphics[width=0.25\linewidth]{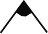}
\end{minipage}
\hspace{-4.5em}
\begin{minipage}{0.9\textwidth}
One ore more sub-determinants can influence LLM-assisted decision-making.
\end{minipage}
\vspace{0,1em}

\begin{minipage}{0.25\textwidth}
  \includegraphics[width=0.25\linewidth]{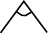}
\end{minipage}
\hspace{-4.5em}
\begin{minipage}{0.9\textwidth}
Exactly one sub-determinant has an effect on LLM-assisted decision-making.
\end{minipage}
\vspace{0,1em} 

\begin{minipage}{0.25\textwidth}
  \includegraphics[width=0.25\textwidth]{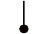}
\end{minipage}
\hspace{-4.5em}
\begin{minipage}{0.9\textwidth}
 Mandatory (sub-)determinants.
\end{minipage}
\vspace{0,1em} 

\begin{minipage}{0.25\textwidth}
  \includegraphics[width=0.25\textwidth]{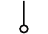}
\end{minipage}
\hspace{-4.5em}
\begin{minipage}{0.9\textwidth}
 Optional (sub-)determinants.
\end{minipage}
\vspace{0,1em} 

\begin{minipage}{0.25\textwidth}
  \includegraphics[width=0.25\textwidth]{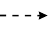}
\end{minipage}
\hspace{-4.5em}
\begin{minipage}{0.9\textwidth}
Interdependencies between determinants.
\end{minipage}
\vspace{0,1em} 


In addition, each determinant is analyzed according to certain comparable aspects, from characteristics and interactions of the determinants in general to implications for LLM-assisted decision-making and scenario-based illustrations of their impact in special. 
To enhance accessibility and comparability for readers, the description of the determinants is organized by employing the following symbols:
\\
\\
\begin{minipage}{0.2\textwidth}
  \includegraphics[width=0.3\linewidth]{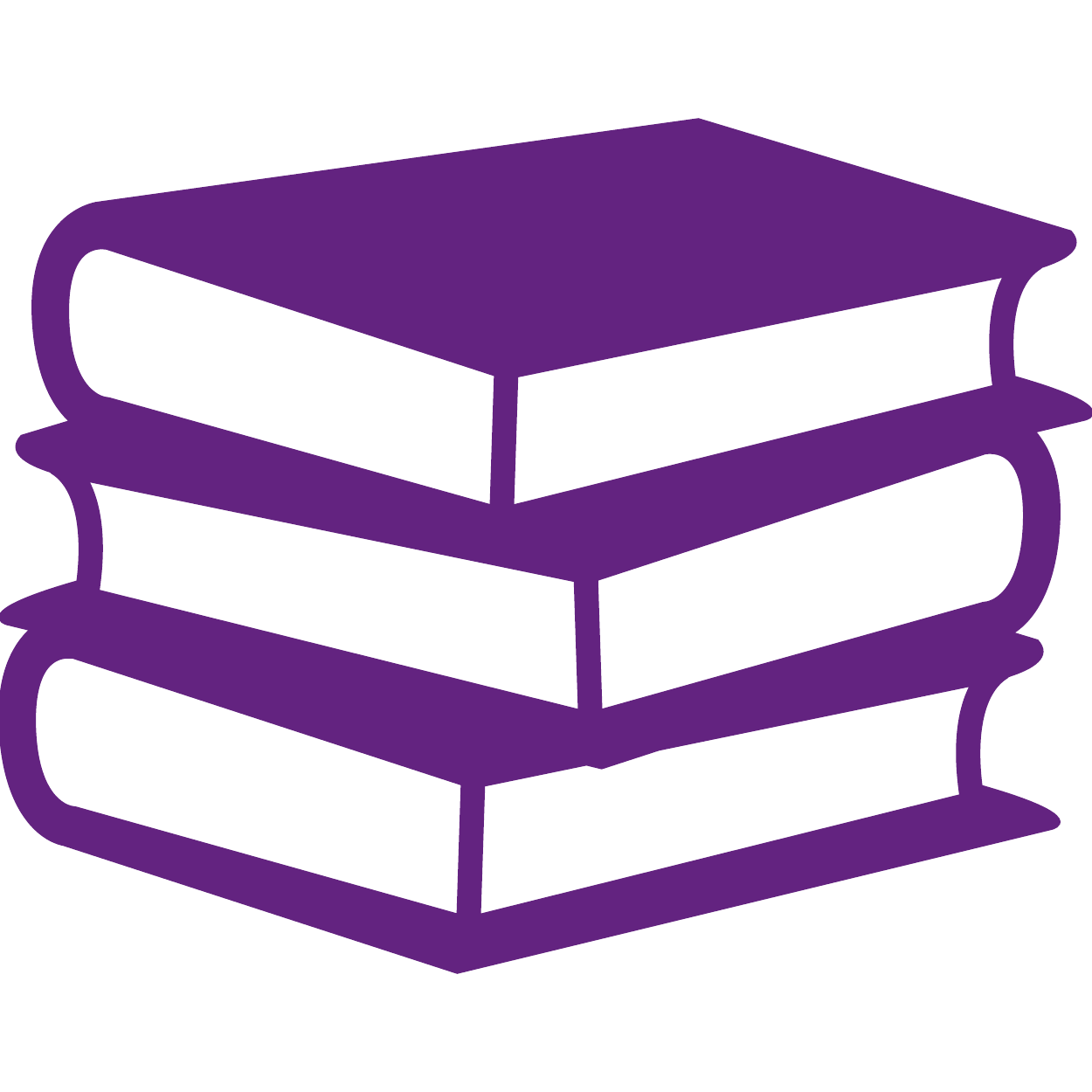}
\end{minipage}
\hspace{-4.5em} 
\begin{minipage}{0,9\textwidth}
 Explanation of determinant's theoretical background.
\end{minipage}

\vspace{0,1em} 
\begin{minipage}{0.2\textwidth}
  \includegraphics[width=0.35\linewidth]{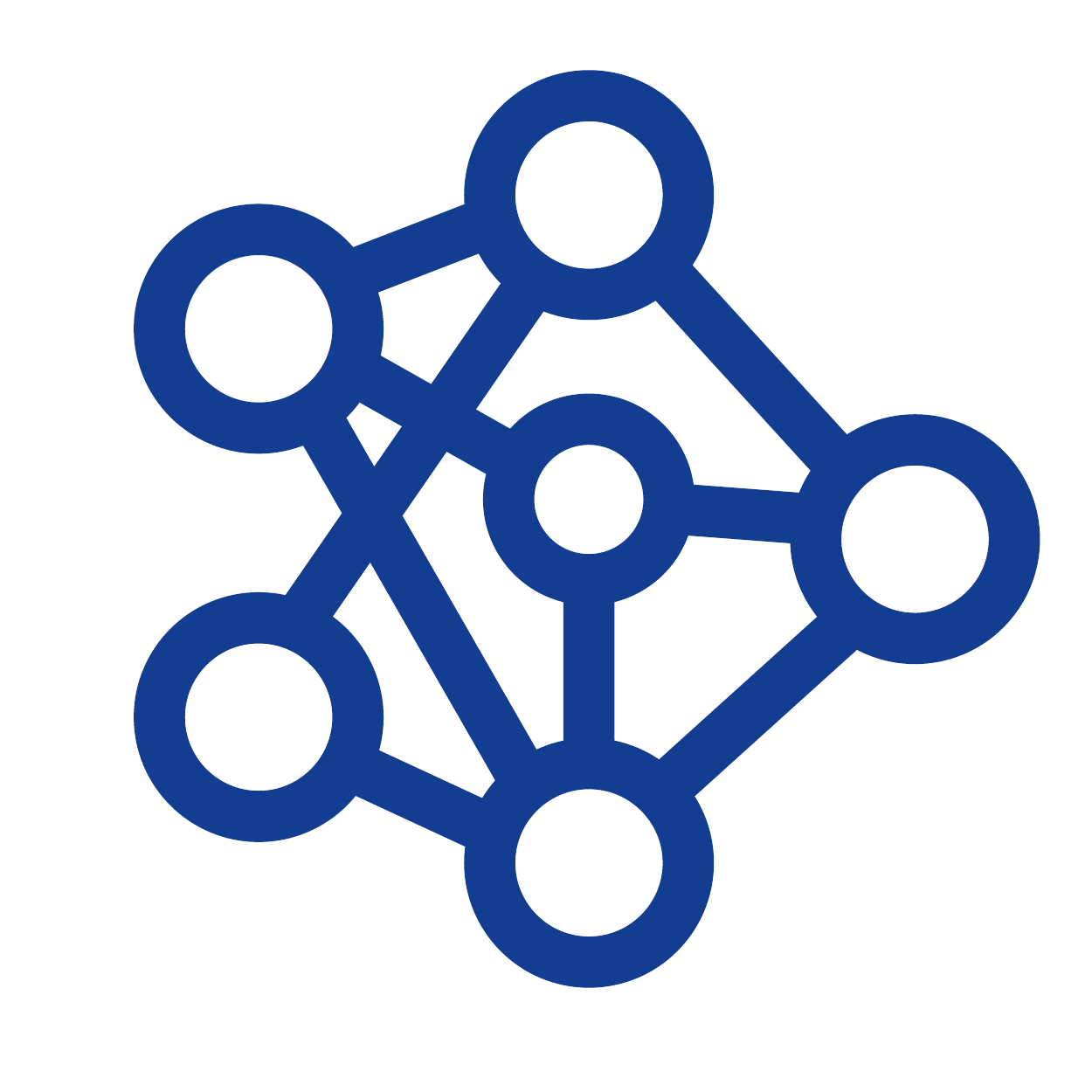}
\end{minipage}
\hspace{-4.5em} 
\begin{minipage}{0.9\textwidth}
  Analysis of interactions with other determinants.
\end{minipage}
\vspace{0,1em} 

\begin{minipage}{0.2\textwidth}
  \includegraphics[width=0.3\linewidth]{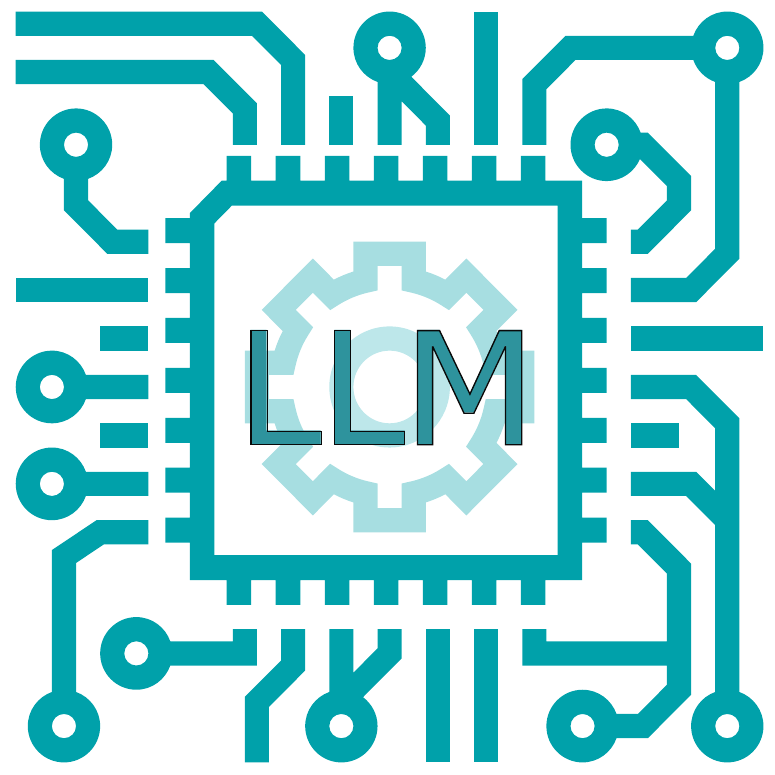}
\end{minipage}
\hspace{-4.5em} 
\begin{minipage}{0.9\textwidth}
  Derivation of implications for LLM-assisted decision-making.
\end{minipage}
\vspace{0,1em} 

\begin{minipage}{0.2\textwidth}
  \includegraphics[width=0.38\textwidth]{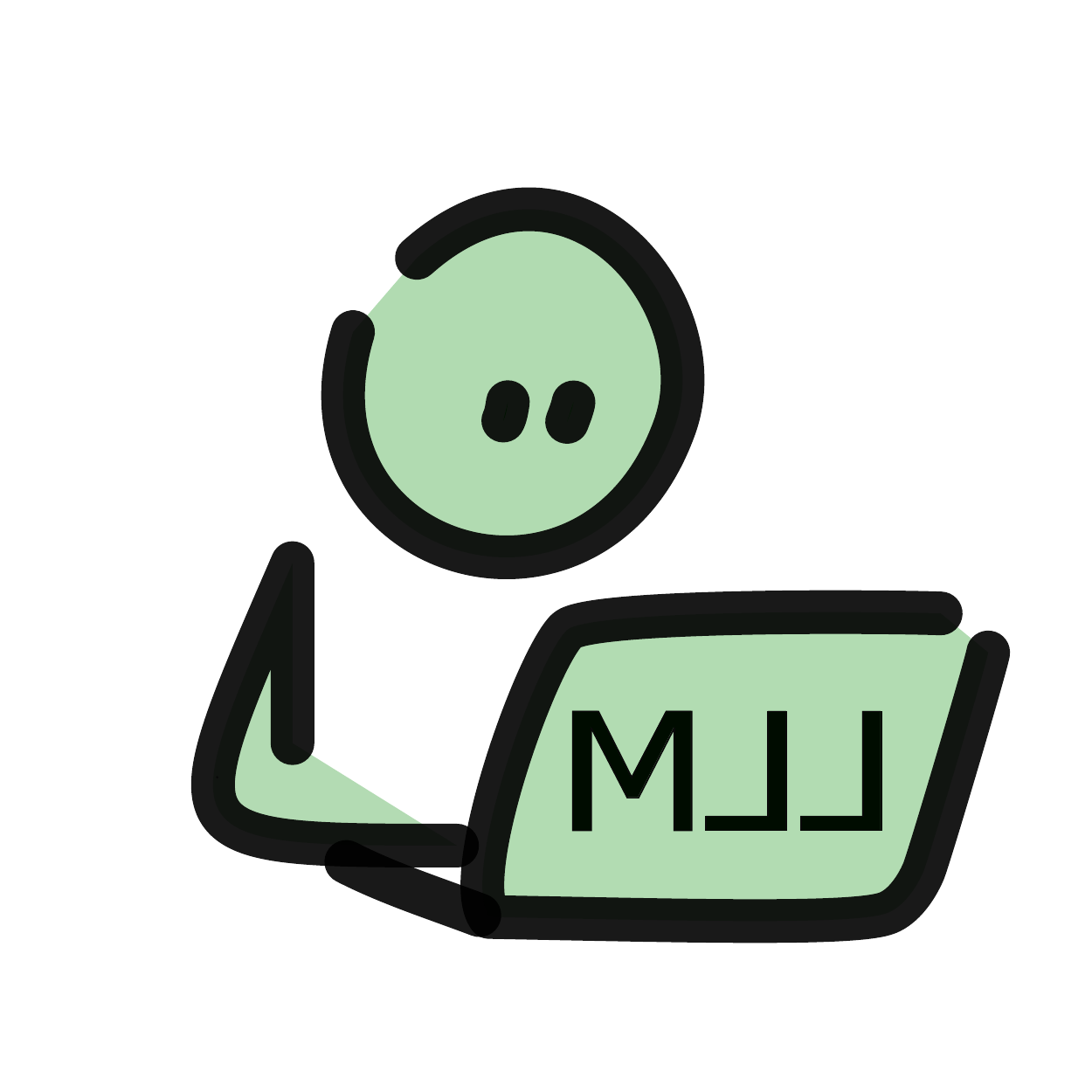}
\end{minipage}
\hspace{-4.5em}
\begin{minipage}{0.9\textwidth}
 Scenario-based illustration of determinant's impact on LLM-assisted decision-making.
\end{minipage}
\vspace{0,1em}

\subsection{Application Scenarios of LLM-assisted Decision-Making}
\label{subsec: Application Scenarios}
Before we explore the determinants of LLM-assisted decision-making and their interactions, we turn our attention to a series of application scenarios to illuminate the significance of understanding these factors. Throughout this paper, we will repeatedly refer to these instances to exemplify the implications and outcomes of the determinants for decisions assisted by LLMs. By dissecting these instances, we aim to provide insights into the effects and interactions of psychological, technological, and decision-specific determinants, and how they can shape the efficacy and efficiency of LLM-assisted decision-making. In the following, the contextual conditions of six illustrative scenarios (S1--S6) are described. 
\begin{compactitem}

\item [\textbf{S1:}] Dr.\ Smith, an experienced physician in a hospital, turned to an LLM-powered medical diagnosis system to identify a patient's symptoms. The system recommended a rare and hard-to-diagnose disease based on the entered symptoms and available medical data. 

\item [\textbf{S2:}] David, a concerned patient, experienced unexplained symptoms and sought answers online. He came across an LLM-powered medical advice forum that, based on his descriptions, suggested a rare illness.

\item [\textbf{S3:}] Anna looked for weight loss advice on the internet. She came across a website, where an LLM provided recommendations. The LLM advised her to follow an extreme diet that eliminated the consumption of almost all carbohydrates and fats. 

\item [\textbf{S4:}] Paula, an expectant mother, turned to an LLM-powered online forum for medical advice regarding her newborn's vaccination. The forum recommended against vaccinating the child based on pseudo-scientific information from unverified sources. 

\item [\textbf{S5:}] Jenna, a marketing manager at a tech startup, sought assistance from an LLM to optimize the company's digital advertising strategy. The LLM suggested investing a significant portion of the marketing budget in a new social media platform that was gaining traction. 

\item [\textbf{S6:}] Alex, a sales manager, turned to an LLM to generate sales projections for the upcoming quarter. The LLM processed historical sales data and market trends to provide detailed sales forecasts. 
\end{compactitem}

In the following Sections \ref{sec:tech-determinants}, \ref{sec:psych-determinants}, and \ref{sec:ds-determinants}, we present the results of our analysis of technological, psychological, and decision-specific determinants of LLM-assisted decision making. 

\section{Technological Determinants of LLM-assisted Decision Making}
\label{sec:tech-determinants}
In this section, we explore technological factors within the realm of LLM-assisted decision-making. 

\begin{figure*}[htb]
	\centering
	\includegraphics[width=1.0\textwidth]{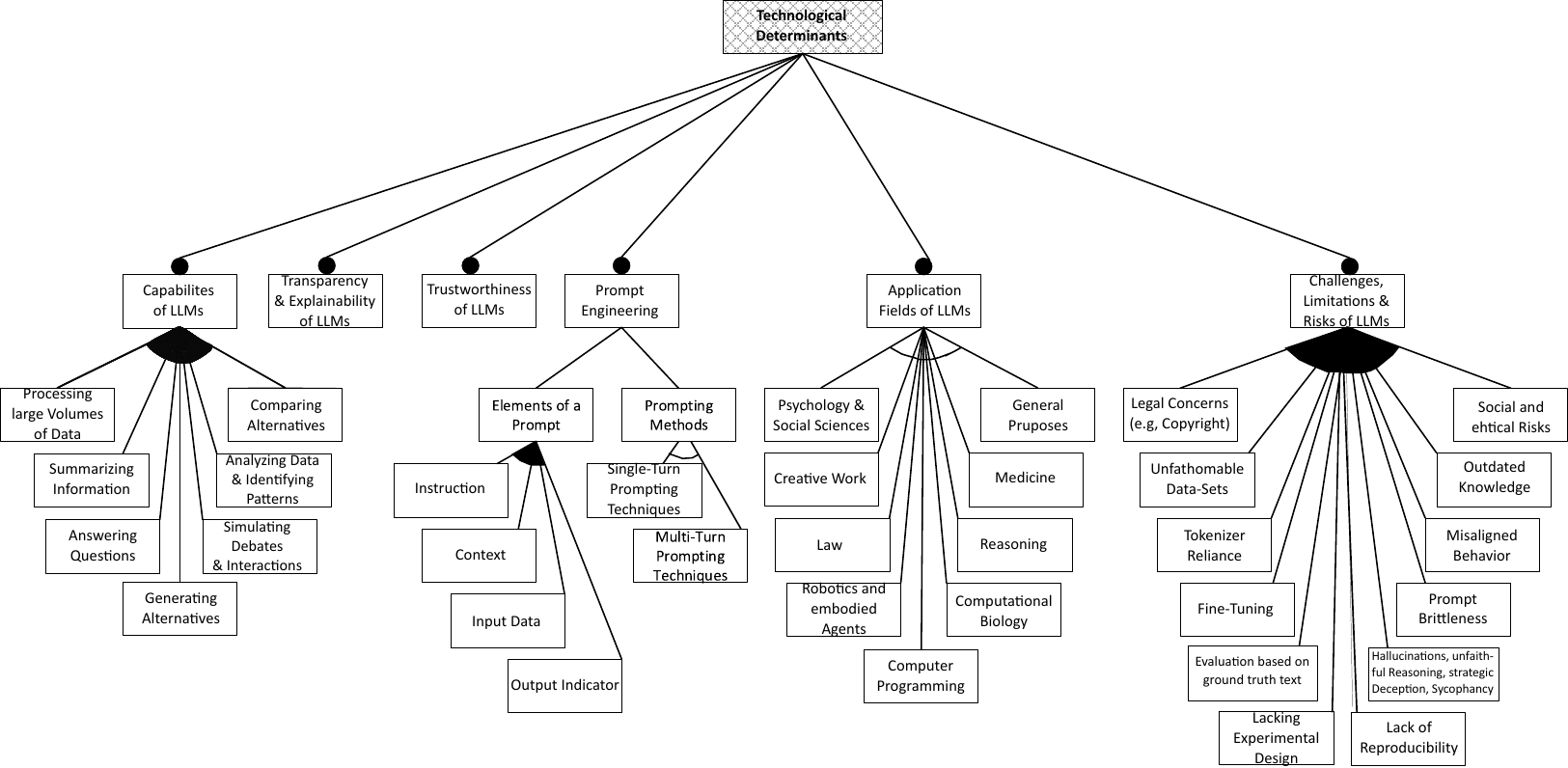}
	\caption{Technological determinants of LLM-assisted decision-making.}
	\label{fig:Technological Determinants of LLM-supported Decision Making}
\end{figure*}

As illustrated by the feature diagram in Figure \ref{fig:Technological Determinants of LLM-supported Decision Making}, technological factors encompass the \texttt{Capabilities of LLMs} (see Section \ref{subsec: Capabilites}), \texttt{Transparency \& Explainability} (see Section \ref{subsec: Transparency}), \texttt{Trustworthiness} (see Section \ref{subsec: Trustworthiness}), \texttt{Prompt Engineering} (see Section \ref{subsec: Prompt Engineering}), \texttt{Application Fields} (see Section \ref{subsec: Application Fields}), as well as \texttt{Challenges, Limitations \& Risks} (see Section \ref{subsec: Challenges}) associated with LLMs. 

As evident from Figure \ref{fig:Technological Determinants of LLM-supported Decision Making}, one or more sub-factors of the determinants \texttt{Capabilities of LLMs} as well as \texttt{Challenges, Limitations \& Risks}, can exert an influence on LLM-assisted decision-making. In \texttt{Prompt Engineering}, both the \texttt{Elements of a Prompt} and the \texttt{Prompting Methods} influence decision processes supported by LLMs. Regarding the \texttt{Elements of a Prompt}, one or more of these components can play a role. In terms of \texttt{Prompting Methods}, either single-turn or multi-turn prompting methods can be applied in decision-making with the support of LLMs. Furthermore, Figure \ref{fig:Technological Determinants of LLM-supported Decision Making} indicates that the use of LLMs is limited to a single \texttt{Application Field} in the decision-making process.

\subsection{Capabilites of LLMs}
\label{subsec: Capabilites}
\begin{minipage}{0.08\textwidth}
  \centering
  \includegraphics[width=0.7\textwidth]{images/Books.pdf}
\end{minipage}%
\begin{minipage}{0.9\textwidth}
  \subsubsection{Theoretical Background}
\end{minipage}

Contrary to smaller language models, which were limited to solving specific problems, LLMs have the ability to tackle a variety of tasks \cite{Chang2023ASO}; also see Section \ref{subsec: LLM}. In particular, LLMs possess the capability to interact and conduct human-like conversations, allowing users to refine their output \cite{Lambert20234ChtGPT}. Scaling has enabled LLMs to achieve state-of-the-art performance in natural language processing (NLP) tasks \cite{Huang2022LargeLM}. Therefore, LLMs are capable of replying to free-text queries without needing specific training for the task at hand \cite{Thirunavukarasu2023LLM}. Moreover, LLMs perform remarkably well in contextual question answering, machine translation and code generation \cite{Qian2023CommunicativeAF}. In addition, LLMs possess reasoning abilities, which serves as a foundation for problem-solving, decision-making and critical analysis that can be triggered, for example, by \textit{Chain-of-Thought (CoT)} prompting \cite{Xi2023TheRA}. By considering different reasoning paths and evaluating decisions to determine the next course of action, LLMs are increasingly capable of deliberate problem solving and decision-making \cite{Yao2023TreeOT}. Owing to their natural language understanding and generation abilities, LLMs can conduct conversations in different languages, enabling individuals to understand and interact with them \cite{scao2022bloom}. Regarding information retrieval, LLMs can, for example, function as a search engine, summarize documents, or interpret texts. Additionally, LLMs are capable of assisting in writing, such as in research or creative writing \cite{Liu2023TrustworthyLA}. Furthermore, LLMs have the ability to generate ideas \cite{Girotra2023Ideas}. LLMs have shown the capability to identify patterns \cite{jin2023time} and analyze historical data \cite{rane2023chatgpt}. Moreover LLMs show the ability to simulate social interactions \cite{park2022social} and debates encompassing various opinions \cite{taubenfeld2024systematic}. Additionally, LLMs display a significant level of rationality in decision-making tasks \cite{chen2023emergence}. 

\begin{minipage}{0.08\textwidth}
  \centering
  \includegraphics[width=0.7\textwidth]{images/LLM1.pdf}
\end{minipage}%
\begin{minipage}{0.9\textwidth}
 \vspace{1em}
  \subsubsection{Deriving Implications for LLM-assisted Decision-Making} 
\end{minipage}
LLMs can be used for different purposes to support human decision-making. As LLMs have the ability to process and summarize large volumes of text data \cite{Liu2023TrustworthyLA}, they can support decision-makers to quickly understand key insights. LLMs have shown the capability to generate ideas \cite{Girotra2023Ideas}. Hence, they can assist in generating alternatives in decision-making processes. As LLMs have demonstrated the ability to adopt personas of different characters \cite{park2022social} and simulate discussions from various viewpoints  \cite{taubenfeld2024systematic}, they enable human decision-makers to consider numerous scenarios and potential outcomes before reaching a decision. Due to their analytical capabilities LLMs, such as identifying patterns \cite{jin2023time} and analyzing historical data \cite{rane2023chatgpt}, LLMs are able to conduct comparative analyses by systematically processing and analyzing available datasets, examining the advantages and disadvantages inherent in each alternative. Such an analytical approach can facilitate a nuanced understanding of the relative strengths and weaknesses associated with diverse choices, thereby providing a data-driven foundation for decision-making. Additionally, LLMs show a considerable capacity for rational decision-making \cite{chen2023emergence} suggesting that they have the potential to improve human decision-making processes through the provision of well-reasoned outputs.

An overview of the \texttt{Capabilities of LLMs} significant for decision-making can be found in Figure \ref{fig:Technological Determinants of LLM-supported Decision Making}.

\subsection{Transparency and Explainability of LLMs}
\label{subsec: Transparency}
\begin{minipage}{0.08\textwidth}
  \centering
  \includegraphics[width=0.7\textwidth]{images/Books.pdf}
\end{minipage}%
\begin{minipage}{0.9\textwidth}
  \subsubsection{Theoretical Background}
\end{minipage}
Black-box models operate as opaque systems, whose internal workings are challenging to access or interpret. These models generate predictions and recommendations based on input data, yet the underlying decision-making process and reasoning remain nontransparent \cite{HassaniTheRole2023}. Due to this opacity in the "inner" working mechanisms of LLMs and their high complexity, they are often categorized as "black-box" models \cite{Zhao2023ExplainabilityFL}. Transparency indicates the extent to which the inherent operational rules and internal logic of a technology are evident to users \cite{Hoff2015Trust}. Explanations can serve as a form of transparency \cite{Zerilli2022howtransparency}. An AI system is considered explainable if it is "intrinsically interpretable or if the non-interpretable task model is complemented with an interpretable and faithful explanation" \cite[p.2]{Markus2021Therole}. Explainability comprises two key aspects: interpretability, which is the extent to which a person can understand an explanation, and fidelity, which refers to the descriptive precision of an explanation \cite{Markus2021Therole}. In many fields, such as medical diagnostics, providing an explanation of how an answer has been generated is essential for fostering transparency and trust. Explanations can enhance user's understanding of black box models \cite{lakkaraju2017interpretable, Ribeiro2018AnchorsHM}, and increase the transparency of AI models \cite{Nourani2021AnchoringBA}. 

LLMs possess the capacity to offer (seemingly) reasonable explanations, called self-explanations. For instance, when tasked with solving a math problem, they frequently present detailed derivation steps, even without explicit instructions to do so. Likewise, in the analysis of the sentiment of a  book review, they spontaneously justify their decisions, providing supporting evidence \cite{huang2023can}. However, LLMs might not consistently convey their "thoughts" accurately, probably decreasing transparency \cite{turpin2023language}.

\begin{minipage}{0.08\textwidth}
  \centering
  \includegraphics[width=0.7\textwidth]{images/LLM1.pdf}
\end{minipage}%
\begin{minipage}{0.9\textwidth}
 \vspace{1em}
  \subsubsection{Deriving Implications for LLM-assisted Decision-Making}
\end{minipage}
Explanations can enhance users' understanding of LLMs, and therefore may increase their transparency. When users understand the rationale behind an LLM's suggestions, they are able to make more informed decisions. This is particularly important in situations where the LLM's recommendations are one of many factors considered in the decision-making process. Understanding the "why" behind an LLM's output allows users to weigh its suggestions appropriately against other considerations. Consequently, transparency and explainability in LLMs can lead to better decision-making. The provision of explanations by LLMs could wield a substantial influence on transparency. Although these self-explanations may initially enhance the perceived transparency in the decision-making process, it is crucial to acknowledge potential limitations since LLMs might not consistently communicate their "thoughts" accurately. 

\begin{minipage}{0.08\textwidth}
  \centering
  \includegraphics[width=0.75\textwidth]{images/Scenario.pdf}
\end{minipage}%
\begin{minipage}{0.9\textwidth}
 \vspace{1em}
  \subsubsection{Scenario-based Demonstration} 
\end{minipage}
\textbf{Transparency and Explainability (Scenario 4).} In the scenario involving Paula, an expectant mother seeking medical advice on an LLM-powered online forum, transparency and explainability are crucial for ensuring informed decision-making. If the forum had been transparent about its sources of information, Paula would have been able to identify the origin of advice, discerning whether it came from verified medical professionals or scientific studies rather than unverified or pseudo-scientific sources. For example, if the forum had clearly stated that its recommendation against vaccination was based on unverified sources or personal opinions, rather than on scientific consensus, Paula might have approached the advice with greater caution. Moreover, if the LLM was capable of explaining the rationale behind its recommendation, including the data and sources it utilized, Paula would have a better understanding of the advice's foundation. If the LLM explained that its recommendation was based on pseudo-scientific information, Paula might recognize the need for further consultation with healthcare professionals. 

\textbf{Transparency and Explainability (Scenario 5).} 
Transparency from the LLM about the data behind its recommendation could have helped Jenna discern whether the new social media platform's popularity was a short-lived trend or had long-term potential. Explainability of the LLM's decision-making process would have allowed Jenna to comprehend the reasoning behind the recommendation, such as the basis of high engagement rates from similar campaigns, and to evaluate their relevance to her company's context.

\subsection{Trustworthiness of LLMs}
\label{subsec: Trustworthiness}

\begin{minipage}{0.08\textwidth}
  \centering
  \includegraphics[width=0.7\textwidth]{images/Books.pdf}
\end{minipage}%
\begin{minipage}{0.9\textwidth}
  \subsubsection{Theoretical Background}
\end{minipage}
The placement of trust in someone often requires a belief in their trustworthiness \cite{Liu2023TrustworthyLA}. Trust and trustworthiness are closely connected yet fundamentally different concepts \cite{hardin2006trust}. Trust is commonly viewed as an attitude, whereas trustworthiness is a deliberate action pertaining to a quality inherent in the object of this attitude, satisfying it and contributing to its adequacy \cite{Nickel2010Can}. Liu et al.\ propose a taxonomy of seven categories influencing trustworthiness of LLMs \cite{Liu2023TrustworthyLA}: In this context, \textit{Reliability} refers to “generating correct, truthful and consistent output with proper confidence”. \textit{Safety} concerns the aspect of preventing unsafe and illegal outputs and the leakage of private information. The \textit{Resistance to Misuse} category includes inhibiting abuse by malicious attackers to cause harm. The \textit{Explainability and Justification} category refers to the ability to explain and correctly justify the outputs to users. The \textit{Social Norm} category concerns the reflection of commonly shared human values. \textit{Robustness} involves the resilience to adversarial attacks and distributional shifts. \textit{Fairness} refers to avoiding bias, not favouring certain groups of users or ideas, and not using stereotypes. \textit{Robustness} refers to the resistance to adversarial attacks and distributional shift.

\begin{minipage}{0.08\textwidth}
  \centering
  \includegraphics[width=0.75\textwidth]{images/Interactions1.pdf}
\end{minipage}%
\begin{minipage}{0.9\textwidth}
 \vspace{1em}
  \subsubsection{Interactions with other Determinants} 
\end{minipage}
\textbf{Transparency and Trustworthiness of LLMs.} Due to the so-called \textit{black box} problem, which refers to the complexity and opacity of the structure, internal working and system implementation \cite{Adadi2018Peeking}, AI systems are becoming increasingly complex, rendering them difficult to understand. This complexity reduces the perceived trustworthiness of the system, as it becomes challenging to find explanations and reasoning for the output \cite{Kaur2022TrustworthyAI}. The transparency of LLMs can be considered a potential determinant of their trustworthiness \cite{Liu2023TrustworthyLA}. 

\begin{minipage}{0.08\textwidth}
  \centering
  \includegraphics[width=0.70\textwidth]{images/LLM1.pdf}
\end{minipage}%
\begin{minipage}{0.9\textwidth}
 \vspace{1em}
  \subsubsection{Deriving Implications for LLM-assisted Decision-Making} 
\end{minipage}
The complexity of LLMs, compounded by the black box problem, implies that understanding their internal workings is challenging. This lack of transparency makes it difficult for users to comprehend how LLMs arrive their suggestions or decisions, thereby reducing the trustworthiness of these models. Hence, transparency significantly influences the trustworthiness of LLMs. When users have a clearer understanding of how LLMs operate and make suggestions, they are more likely to trust the outcomes, leading to an enhanced trustworthiness of LLMs. 

\begin{minipage}{0.08\textwidth}
  \centering
  \includegraphics[width=0.75\textwidth]{images/Scenario.pdf}
\end{minipage}%
\begin{minipage}{0.9\textwidth}
 \vspace{1em}
  \subsubsection{Scenario-based Demonstration} 
\end{minipage}
\textbf{Trustworthiness and Transparency (Scenario 3).} In the example of Anna seeking weight-loss advice from an LLM, enhanced trustworthiness through transparency might have influenced her decision-making process. If the LLM's recommendations has been transparent, clearly detailing the sources and data used to formulate the advice, Anna could have made a more informed decision. For instance, if the LLM transparently cites reputable nutritional studies or guidelines that support its extreme diet recommendation, Anna can trust that the advice is grounded in scientific evidence rather than being arbitrary suggestions. The trustworthiness of the LLM can be further enhanced if it is transparent about its limitations and advises Anna to consult a healthcare professional before making drastic dietary changes. Trustworthiness would be increased if the LLM not only provides recommendations but also informs about potential risks, especially concerning extreme diets. 
\vspace{1.5em}

\subsection{Prompt Engineering}
\label{subsec: Prompt Engineering}

\begin{minipage}{0.08\textwidth}
  \centering
  \includegraphics[width=0.7\textwidth]{images/Books.pdf}
\end{minipage}%
\begin{minipage}{0.9\textwidth}
  \subsubsection{Theoretical Background}
\end{minipage}

A prompt consists of a set of instructions that adjust the LLM and/or enhance or refine its capabilities. A prompt establishes the context of the conversation and informs the LLM about which information is crucial, as well as the preferred form of output and desired content \cite{White2023Aprompt}. Prompt engineering refers to the "practice of designing, refining, and implementing prompts or instructions that guide the output of LLMs to help in various tasks." \cite[p.1]{Mesk2023Prompt}. Prompting enhances the efficient utilization of LLMs across various applications and research domains \cite{White2023Aprompt}. The elements of a prompt are the following \cite{Giray2023PromptEW}:

\begin{compactitem}
    \item [1.]\texttt{Instruction}: A specific task that guides the model’s behavior toward the intended output.
    \item [2.]\texttt{Context}: External information that provides background knowledge to the model, enhancing the accuracy and relevance of its responses.
    \item [3.]\texttt{Input Data}: The query or information that requires the model's processing and response, forming the core of the prompt.
    \item [4.]\texttt{Output Indicator}: A definition of the desired response format, such as a brief answer, a paragraph, or any other specific layout, which shapes the model’s reply accordingly.
\end{compactitem}

Understanding the elements of a prompt is essential as it enables users to clearly convey their intentions to the model, and, hence, to guide the model's behavior and effectively enhance the quality of its responses \cite{Mesk2023Prompt}.

Karmaker and Teler \cite{Karmaker2023TELeRAG} propose a taxonomy to categorize LLM prompts for complex tasks based on the following four dimensions: 
\begin{compactitem}
    \item [1.]\textit{Turn}: This dimension refers to the number of turns applied while prompting an LLM.
    \item [2.]\textit{Expression}: Depending on how the task and its sub-tasks are articulated, prompts can be categorized as either question-style or instruction-style.
    \item [3.]\textit{Role}: This dimension classifies prompts based on whether a specific system role is defined in the LLM system prior to presenting the actual prompt. Prompts may have either a defined or undefined system role.
    \item [4.]\textit{Level of Details}: Prompts are categorized based on the presence or absence of specific elements of the goal task definition in the instruction. 
\end{compactitem}

Accordingly, prompting methods can be divided into \texttt{Single-Turn} and \texttt{Multi-Turn Prompting Techniques} \cite{Karmaker2023TELeRAG}. Single-Turn Prompting methods involve prompts that elicit an answer in one shot, while Multiple-Turn Prompting Techniques iteratively chain prompts and their responses \cite{kaddour2023challenges}. A prominent example of Single-Turn Prompting is \textit{Chain-of-Thought} prompting, which decomposes a multi-step problem into intermediate steps \cite{Wei2022ChainOT}. An example of Multi-Turn Prompting Techniques is the \textit{Tree-of-Thoughts}. This approach extends the \textit{Chain of Thought} to obtain a tree of thoughts with multiple distinct paths, where each thought serves as an intermediate step. The \textit{Tree-of-Thoughts} enables the LLM to self-assess the progress of these intermediate thoughts and incorporate search algorithms that systematically explore of the tree \cite{Yao2023TreeOT}.

\begin{minipage}{0.08\textwidth}
  \centering
  \includegraphics[width=0.75\textwidth]{images/Interactions1.pdf}
\end{minipage}%
\begin{minipage}{0.9\textwidth}
 \vspace{1em}
  \subsubsection{Interactions with other Determinants} 
\end{minipage}
\textbf{Prompt Engineering and Transparency of LLMs.} Prompting can enhance transparency in several ways. Chaining significantly improves the quality of system transparency. In a chain, a problem is divided into several smaller sub-tasks. Each of these is associated with a distinct step, accompanied by a specific natural language prompt. The outcomes of one or more preceding steps are then compiled and included in the input prompt for the subsequent step \cite{Wu2022AI}. Moreover, users can gain insights into how the model arrives at its conclusion, as prompting can lead the LLM to reveal its thought processes \cite{Wei2022ChainOT}. By asking the LLM to provide analogies via prompting \cite{bhavya2022analogy} responses of LLMs can become more understandable and transparent. 
\\
\\
\textbf{Prompt Engineering and Capabilities of LLMs.} Several prompting strategies have been developed with the aim of enhancing reasoning and compositional capabilities \cite{chen2023skills}. For instance, \textit{Chain-of-Thought} prompting \cite{Wei2022ChainOT} enhances reasoning performance of LLMs by illustrating how a problem can be addressed through a sequence of simple steps. Another example is \textit{Skills-in-Context} prompting, which enhances question answering, dynamic programming, and math reasoning. \textit{Skills-in-Context} prompting comprises two fundamental components: the foundational skills essential for problem-solving and examples illustrating how to integrate these skills into solutions for intricate problems \cite{chen2023skills}.

\begin{minipage}{0.08\textwidth}
  \centering
  \includegraphics[width=0.75\textwidth]{images/Scenario.pdf}
\end{minipage}%
\begin{minipage}{0.9\textwidth}
 \vspace{1em}
  \subsubsection{Scenario-based Demonstration} 
\end{minipage}
\textbf{Prompt Engineering and Transparency of LLMs (Scenario 6).} In the example of Alex, a sales manager using an LLM to generate sales projections, transparency induced by effective prompting can significantly influence the output and, consequently, the decision-making process. By using prompts that request the LLM to explain how it arrived at its sales projections, Alex can gain insights into the factors the model considered. This might involve an analysis of how historical sales data and current market trends were evaluated and factored into the forecast. Such transparency can help Alex comprehend the rationale behind the projections.

\textbf{Prompt Engineering and Capabilities of LLMs (Scenario 1).} 
By using Single-Turn Prompting, Dr.\ Smith could promptly receive an initial recommendation for a rare disease based on the entered symptoms. However, the system's comprehension is confined to the information presented in that single prompt, potentially overlooking nuances or changes in the patient's condition. In contrast, through Multi-Turn Prompting, Dr.\ Smith can provide additional details, request more specific information, or seek clarifications. This iterative process helps the LLM consider a broader context, leading to more nuanced and accurate recommendations, especially for rare and hard-to-diagnose diseases.
\vspace{1.5em}

\subsection{Application Fields of LLMs}
\label{subsec: Application Fields}
\begin{minipage}{0.08\textwidth}
  \centering
  \includegraphics[width=0.7\textwidth]{images/Books.pdf}
\end{minipage}%
\begin{minipage}{0.9\textwidth}
  \subsubsection{Theoretical Background}
\end{minipage}
The potential applications of LLMs are diverse. These applications can vary from general uses, such as chatbots that incorporate functions of information acquisition, multi-turn interaction, and text generation, to more specific purposes. Currently, LLMs are being applied in various domains, including the following \cite{kaddour2023challenges}:
\begin{compactitem}
    \item Chatbots for \texttt{General Purposes}, which integrate functions of information acquisition, multi-turn interaction, and text generation,
    \item \texttt{Computational Biology}, such as protein embeddings or genomics analyses,
    \item \texttt{Computer Programming}, for example code generation,
    \item Creative work, primarily applied for story and script generation,
    \item \texttt{Law}, such as answering legal questions, finding related precedents, and generating legal text,
    \item \texttt{Medicine}, for example answering medical questions or retrieving medical information,
    \item \texttt{Reasoning}, e.g., for mathematical or arithmetical tasks,
    \item \texttt{Robotics and embodied Agents}, e.g., for providing high-level planning and contextual knowledge, and
    \item \texttt{Psychology and Social Sciences}, e.g., for modeling human behavior or analyzing behavioral characteristics of LLMs.
\end{compactitem}

Possible application fields of LLMs within human decision-making processes are illustrated in Figure \ref{fig:Technological Determinants of LLM-supported Decision Making}.

\subsection{Challenges, Limitations \& Risks of LLMs}
\label{subsec: Challenges}
\begin{minipage}{0.09\textwidth}
  \centering
  \includegraphics[width=0.7\textwidth]{images/Books.pdf}
\end{minipage}%
\begin{minipage}{0.9\textwidth}
  \subsubsection{Theoretical Background}
\end{minipage}

Challenges associated with decisions made prior to the implementation of an LLM, include, for example, \texttt{Unfathomable Datasets}, \texttt{Fine-Tuning}, and \texttt{Tokenizer-Reliance}. \textit{Unfathomable Datasets} refer to the issue that the size of the pre-training datasets currently in use is so large that it becomes nearly impossible for individuals to validate the quality of the documents they contain \cite{kaddour2023challenges}. For instance, the dataset of LLMs contains numerous near-duplicates, which negatively impacts the models' performance \cite{lee2021deduplicating}. An additional obstacle involves \textit{Fine-Tuning} required for integrating up-to-date item information, which, in turn, demands substantial computational resources and incurs time costs \cite{lin2024data}. Challenges related to Tokenizers include computational overhead, dealing with new words, and low interpretability on the user side \cite{kaddour2023challenges}. Despite their capabilities, LLMs are susceptible to errors, particularly if they have been trained on biased or incomplete data. Given their continuous learning from internet texts, neglecting to thoroughly verify and validate LLMs' responses may lead to incorrect or incomplete decisions \cite{choudhury2023investigating}.

Behavioral challenges of LLMs that emerge during deployment include \texttt{Prompt Brittleness}, \texttt{Misaligned Behavior} and \texttt{Outdated Knowledge}. \textit{Prompt Brittleness} \cite{parameswaran2023revisiting} refers to the phenomenon where even  modifications in wording can significantly impact the overall accuracy \cite{Zamfirescu2023Why}. \textit {Misaligned Behavior} points to the fact that outputs generated by LLMs often do not align well with human values or intentions, resulting in unintended or adverse consequences \cite{Gabriel2020Artificial, russell2019human}. Moreover, the knowledge incorporated into LLMs might become outdated or inappropriate as time progresses \cite{wu2023eva}. Another limitation of LLMs is that the high complexity and scaling of LLMs pose challenges in terms of explainability \cite{gao2023chat, Hadi2023Large}. 

LLMs encounter an additional limitation known as \texttt{Hallucinations}, where these models generate information that appears plausible but is factually incorrect, including the fabrication of non-existent facts \cite{xu2024hallucination, yao2023llm}. Further problematic behaviors of LLMs include mirroring the viewpoints introduced by users (\texttt{Sycophancy}) \cite{perez2022discovering}, offering rationalizations that do not align with the actual reasons behind the LLMs' outputs (\texttt{Unfaithful Reasoning}) \cite{turpin2023language}, and engaging in deception when LLMs deduce that it could further a specific goal (\texttt{Strategic Deception}) \cite{park2023ai}. Additionally, LLMs can pose \texttt{Social \& Ethical Risks}. For example, LLMs can perpetuate unfair discrimination through stereotyping and social prejudice. Risks also arise from the potential leakage of private data or the ability of LLMs to correctly infer private or sensitive information. Other threats relate to the use of LLMs for harmful purposes, such as fraud or the development of computer code for viruses. Additional risks steem from the perception of the system as "human-like," which may lead users to overestimate its capabilities, resulting in over-reliance or unsafe use \cite{Weidinger2021EthicalAS}.

Challenges that hinder academic progress include \texttt{Evaluations based on ground truth Text} written by individuals, \texttt{Lacking experimental Design}, such as ablations, and \texttt the {Lack of Reproducibility} in LLM research \cite{kaddour2023challenges}.

\begin{minipage}{0.08\textwidth}
  \centering
  \includegraphics[width=0.75\textwidth]{images/Interactions1.pdf}
\end{minipage}%
\begin{minipage}{0.9\textwidth}
 \vspace{1em}
  \subsubsection{Interactions with other Determinants} 
\end{minipage}
\textbf{Limitations, Transparency, and Explainability of LLMs.} Research has indicated that LLMs might not consistently convey their "thoughts" accurately, as, for instance, \textit{Chain-of-Thought} explanations have the potential to systematically misrepresent the actual basis for a model's prediction, potentially negatively impacting transparency. Hence, it can't be assumed by default that explanations provided by LLMs are faithful, meaning they represent the actual reasons behind the model's predictions \cite{turpin2023language}. One of the reasons, for instance, is that human-crafted explanations integrated into the training of LLMs are incomplete, frequently excluding significant parts of information of the causal chain leading to an outcome \cite{Lombroze2006The,turpin2023language}.  

Figure \ref{fig:Technological Determinants of LLM-supported Decision Making} provides an overview of \texttt{Challenges, Limitations \& Risks} among other aspects.
\\
\\
\vspace{-2mm}
\section{Psychological Determinants of LLM-assisted Decision-Making}
\label{sec:psych-determinants}
\begin{figure*}[htb]
	\centering
	\includegraphics[width=1.0\textwidth]{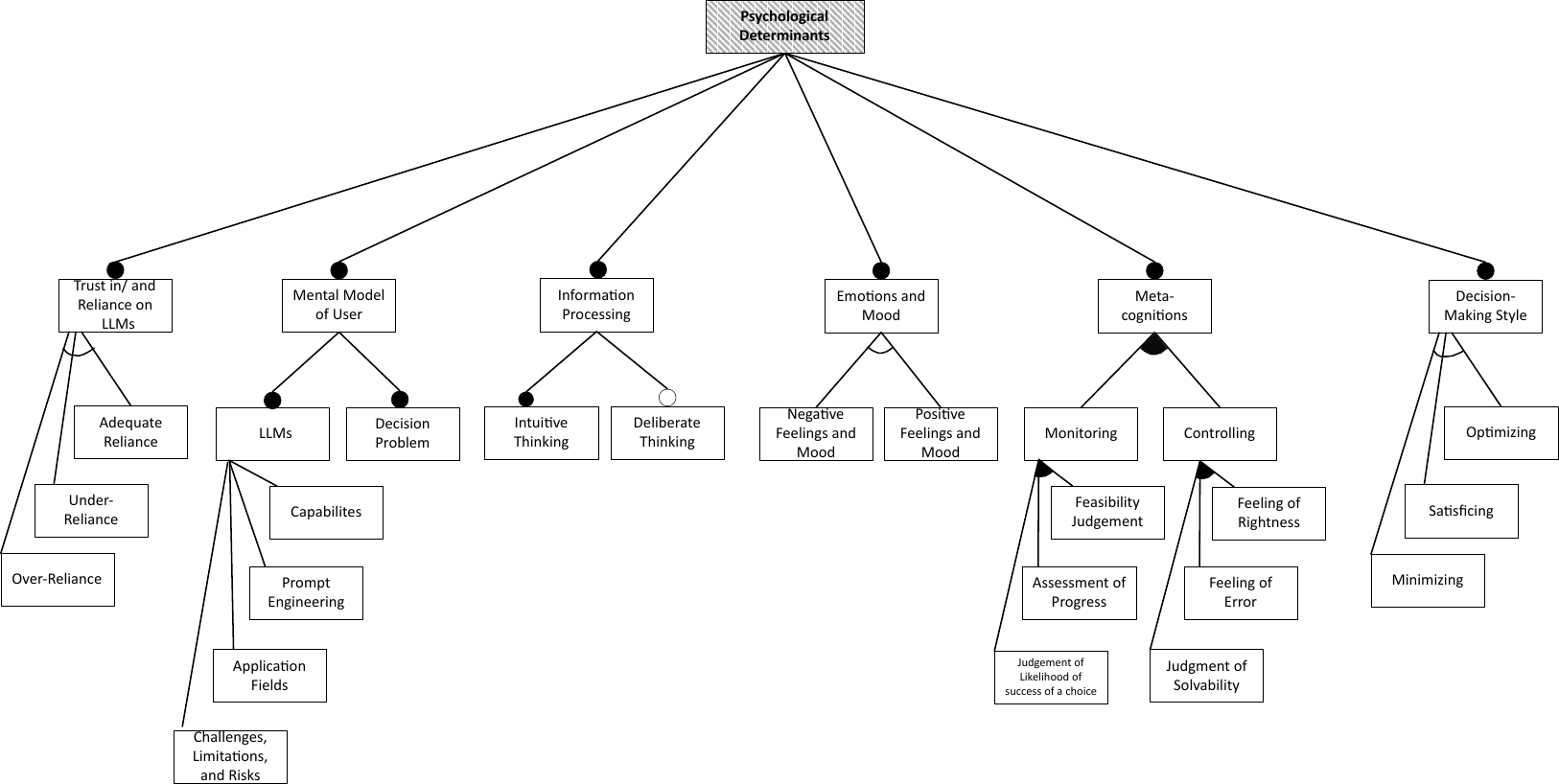}
	\caption{Psychological determinants of LLM-assisted decision-making.}
	\label{fig:psych-determinants}
\end{figure*}

\vspace{-1mm}
This section addresses specific psychological determinants in the context of LLM-assisted decision-making, which are illustrated by the feature diagram in Figure \ref{fig:psych-determinants}. 

As illustrated in Figure \ref{fig:psych-determinants}, \texttt{Trust in/ and Reliance on LLMs} (see Section \ref{subsec:trustReliance}), the \texttt{Mental Model of User} (see Section \ref{subsec: Mental Model}), \texttt{Information Processing} (see Section \ref{subsec: Information Processing}), \texttt{Emotions and Mood} (see Section \ref{subsec: Emotions and Mood}), \texttt{Metacognitions} (see Section \ref{subsec: Metacognitions}), and \texttt{Decision-Making Styles} (see Section \ref{subsec: Decision-making styles}) are mandatory psychological determinants in the LLM-assisted decision-making process. 

Concerning the determinant Trust in/ and Reliance on LLMs, only one of the mentioned sub-determinants \texttt{Adequate Reliance}, \texttt{Under-Reliance} or \texttt{Over-Reliance}, can have an impact on the decision-making process. Figure \ref{fig:psych-determinants} also shows that both the user's mental model regarding LLMs and the decision problem influence LLM-assisted decision-making. In information processing, \texttt{Intuitive Thinking} always influences the decision-making process, whereas \texttt{Deliberate Thinking} may not necessarily occur. As can be inferred from Figure \ref{fig:psych-determinants}, either \texttt{Positive} or \texttt{Negative Feelings and Mood} have an impact on LLM-assisted decision-making. Regarding \texttt{Metacognitions}, \texttt{Monitoring} and/ or \texttt{Controlling} can occur. Of the sub-determinants, controlling and monitoring, either one or more can influence the decision-making process. Furthermore, Figure \ref{fig:psych-determinants} illustrates that only one of the three \texttt{Decision-Making Styles}, i.e., \texttt{Optimizing}, \texttt{Satisficing} or \texttt{Minimizing}, can exert an impact on decision processes assisted by LLMs.

\vspace{-2mm}
\subsection{Trust in/ Reliance on LLMs}
\label{subsec:trustReliance}
\begin{minipage}{0.08\textwidth}
  \centering
  \includegraphics[width=0.7\textwidth]{images/Books.pdf}
\end{minipage}%
\begin{minipage}{0.9\textwidth}
  \subsubsection{Theoretical Background}
\end{minipage}
\texttt{Trust} is a significant determinant in both the implementation and the utilization of AI systems \cite{Lukyanenko2022Trust}. Therefore, trust plays a pivotal role in the interaction between humans and AI, as an insufficient level of trust can result in the misuse or avoidance of the technology \cite{Jacovi2020FormalizingTI}. Trust can be defined as “the reliance by an agent that actions prejudicial to their well-being will not be undertaken by influential others” \cite[p.24]{Hancock2011Can}. In AI-assisted decision-making, individuals need to discern when to trust the AI and when to trust themselves \cite{Shuai2023Who}. Users require a sufficient understanding of LLM applications to control the application's behavior and establish an appropriate level of trust \cite{Liao2023AITI}. 

Achieving complementary performance in AI-assisted decision-making depends on adequately calibrating the degree of human reliance on AI \cite{Cao2022UnderstandingUR}. While trust is defined as an attitude, reliance is considered a behavior influenced by trust \cite{Lee2004TrustIA}. Reliance with regard to AI models is understood as “user’s behavior that follows from the advice of the system” \cite[p.61]{Scharowski2022TrustAR}. Trust plays a crucial role in determining how much people rely on AI. When trust is low in a highly capable technology, it might result in non-utilization, leading to substantial costs in terms of time and work efficiency. Conversely, excessive trust in technology with limited capabilities can result in misuse \cite{Glikson2020Human}. When considering a prediction or suggestion from an AI system, a human decision-maker can either accept or reject it. \texttt{Adequate Reliance} occurs when a person accepts a correct AI prediction or rejects an incorrect one. \texttt{Under-reliance} occurs when a person rejects a correct AI prediction. \texttt{Over-reliance} occurs when humans fail to correct a wrong AI prediction \cite{Vasconcelos2022ExplanationsCR}. 

Hence, over-reliance refers to the frequency with which individuals agree with AI suggestions, even when they are incorrect \cite{Buccinca2021ToTO}. Over-reliance is one of the most common error types in human-AI decision-making \cite{Buccinca2021ToTO, Zhang2020EffectOC}. The tendency of individuals to overestimate the performance of AI agents compared to their own may lead to an overestimation of the AI agent \cite{Kelly2023CapturingHM}. Insufficient trust in AI can lead to under-reliance on AI, while excessive trust may result in over-reliance \cite{Buccinca2021ToTO, Kaur2020Interpreting, Mohseni2020MachineLE}.

\begin{minipage}{0.08\textwidth}
  \centering
  \includegraphics[width=0.75\textwidth]{images/Interactions1.pdf}
\end{minipage}%
\begin{minipage}{0.9\textwidth}
 \vspace{1em}
  \subsubsection{Interactions with other Determinants} 
\end{minipage}
\textbf{Trust in/ Reliance on AI-Systems and Transparency}. Numerous studies have investigated the correlation between AI system transparency and human trust in AI. For example, a meta-analysis by Kaplan et al. \cite{Kaplan2023Trust} indicated that transparency was positively correlated with trust in AI. Additionally, objective transparency of AI can enhance users’s trust in these systems \cite{Grotenhermen2020Arewe, Hoeddinghaus2020TheAutomation}. In contrast, a lack of system transparency can lead to insufficient trust in the system, which in turn can be a barrier to delegating tasks or decisions to intelligent systems \cite{Shin2020HowDU, Shin2021TheEO}. Moreover, research has indicated that perceived transparency of AI-systems can prompt trust by increasing perceived effectiveness and, at the same time, may reduce trust by enhancing discomfort \cite{LIangru2022AI}. 

\vspace{0,5em}
\begin{minipage}{0.08\textwidth}
  \centering
  \includegraphics[width=0.7\textwidth]{images/LLM1.pdf}
\end{minipage}%
\begin{minipage}{0.9\textwidth}
 \vspace{1em}
  \subsubsection{Deriving Implications for LLM-assisted Decision-Making} 
\end{minipage}
In the context of LLM-assisted decision-making, reliance refers to the degree to which a user depends on the outputs and follows the recommendations provided by the LLM. Appropriate reliance occurs when the user correctly accepts or rejects the suggestions of the LLM, leading to an optimal use of LLMs and, hence, enhancing the user's decision-making process. Over-reliance happens when the user places too much trust in the LLM, accepting its suggestions without sufficient scrutiny. This can lead to a lack of critical evaluation of the LLM's outputs, potentially resulting in decisions that fail to consider important factors the LLM might have missed. Under-reliance occurs when the user does not trust the LLM's outputs enough, even when they are accurate, eventually resulting in the under-utilization of the LLM and potentially leading to less informed or sub-optimal decision outcomes.
\\
\\
\textbf{Trust in and Transparency of LLMs.} Transparency emerges as a crucial determinant in influencing trust in LLM-assisted decision-making. When users possess a clear understanding of how the LLM operates and makes recommendations and decisions, it is likely to instill confidence in the technology. A lack of transparency in an LLM may result in insufficient trust. Without trust, users may hesitate to delegate tasks or decisions to LLMs. However, the transparency of LLMs might have a dual impact on trust. On the one hand, transparency may increase trust by enhancing the perceived effectiveness of the system. On the other hand, heightened transparency can also lead to discomfort, potentially diminishing trust in LLMs.

\vspace{0.5em}
\begin{minipage}{0.08\textwidth}
  \centering
  \includegraphics[width=0.75\textwidth]{images/Scenario.pdf}
\end{minipage}%
\begin{minipage}{0.9\textwidth}
 \vspace{1em}
  \subsubsection{Scenario-based Demonstration} 
\end{minipage}
\textbf{Over-reliance (Scenario 1).} Dr.\ Smith, an experienced physician in a hospital, turned to an LLM-powered medical diagnosis system to identify a patient's symptoms. The system recommended a rare and hard-to-diagnose disease based on the entered symptoms and available medical data. Dr.\ Smith trusted the system's recommendation and made the diagnosis accordingly. As a result, the patient underwent costly and invasive tests and treatments. However, it became apparent over time that the patient's symptoms were not caused by the suspected rare disease. The actual cause was a much more common and easily treatable condition that had not been considered before. Despite Dr.\ Smith's experience, he deferred to the LLM system's diagnosis without applying his own medical judgment or considering differential diagnoses. This suggests an over-reliance on the LLM's output over his own expertise. Moreover, Dr.\ Smith did not verify the LLM's recommendation against other diagnostic possibilities or seek a second opinion, which is a standard practice in medicine when faced with rare or unusual diagnoses.
\\
\\
\textbf{Over-reliance (Scenario 3).} Anna was searching for weight loss advice on the internet and came across a website where an LLM provided recommendations. The LLM advised her to follow an extreme diet that eliminated almost all carbohydrates and fats. Trusting the LLM as a reliable source of information, Anna decided to follow this advice without conducting further research. After a few weeks, she noticed her health deteriorating, feeling weak and tired, and even experiencing hair loss. Concerned about these symptoms, she eventually consulted a doctor, who diagnosed her with a deficiency in essential nutrients due to her extreme diet and warned her about the risks of such radical dietary changes. Anna's trust in the LLM's guidance for an extreme diet, without seeking additional information or professional advice, demonstrates her over-reliance on the technology, neglecting the complexity of human health and nutrition. Her decision to adopt a drastic diet based solely on the LLM's recommendation reflects a belief in the LLM's capabilities as comparable to those of a healthcare professional. This case underscores the importance for individuals to seek information from reliable, scientifically-backed sources, especially regarding medical decisions, and to consult healthcare professionals for informed, personalized medical advice.
\\
\\
\textbf{Under-reliance (Scenario 5)}. Jenna, a marketing manager at a tech startup, utilized an LLM to optimize the company's digital advertising strategy. The LLM suggested investing a significant portion of the marketing budget in a new social media platform that was gaining traction. However, Jenna, being skeptical about the LLM's recommendations, decided to stick with the proven channels the company had traditionally used. Despite the LLM's suggestion, she maintained the current budget allocation to established platforms and refrained from experimenting with the new one. As time passed, the campaign on the established platforms continued to yield steady, but not exceptional, results. The consequences of under-reliance became apparent when competitors embraced innovative platforms and witnessed considerable success. The startup, by not adapting to emerging trends, missed potential opportunities for growth and failed to reach a broader audience.
\\
\\
\textbf{Trust in and Transparency of LLMs (Scenario 6)}. In the context of sales projections, transparency in the LLM's decision-making process might directly influence Alex's trust. The more transparent the LLM is about its data processing, confidence levels, rationale, and assumptions, the more likely Alex is to trust the generated sales forecasts. Transparency regarding any assumptions made by the LLM can help Alex understand the limitations and potential risks associated with the projections. Moreover, clear articulation of the rationale behind each sales projection may build trust. When Alex can see a logical and data-driven basis for the forecasts, it might enhance confidence in the LLM's ability to generate meaningful and accurate predictions.

\vspace{1,5em}
\subsection{Mental Model}
\label{subsec: Mental Model}
\vspace{-2mm}

\begin{minipage}{0.09\textwidth}
  \centering
  \includegraphics[width=0.7\textwidth]{images/Books.pdf}
\end{minipage}%
\begin{minipage}{0.9\textwidth}
  \subsubsection{Theoretical Background} 
\end{minipage}
A crucial factor influencing the efficient use of AI in decision-making is the individual's \texttt{Mental Model of the Decision Problem} and \texttt{the LLM}. Mental models can be regarded as simplified cognitive representations or knowledge structures about how particular aspects of the world work \cite{Gary2016MentalModels}. More precisely "mental models are comprised of interrelated memories, conceptual knowledge, and causal beliefs that create an understanding of how something works in the real world and form expectations about future events" \cite[p.~2]{holtrop2021importance}. Mental models are based on an individual's knowledge, experiences, values, beliefs, expectations, and aspirations, and they explain how people selectively filter, process, and interpret information \cite{EASTERBYSMITH19803}. Therefore, mental models influence learning, problem-solving, and decision-making \cite{Johnson-LairdMentalModels1986, RehderCategorization2003}. Mental models shape the perception of the decision-making system and its elements, including the problem, the decision-making process, the decision outcome, and the feedback \cite{forrester1999industrial, ChermackMentalModels2003}. They comprise individual perceptions of both external and internal variables, choice solutions, decision-making assumptions, and biases \cite{ChermackMentalModels2003}. 

Individuals generate mental models for any system they interact with \cite{NormanPsychology1988}. This also applies to AI agents \cite{KuleszaTellmemore2012}. Such mental models encompass a persons' beliefs about the AI and their expectations regarding the outcomes of interacting with it \cite{Steyvers2023Three}. Hence, users potentially hold preconceived expectations during the interaction with AI systems \cite{GrimesMentalModels2021}. In scenarios where the human is responsible for deciding when and how to use the AI system's recommendation, they need to build knowledge, i.e., mental models, of the AI's different capabilities \cite{Bansal2019BeyondAT}. Through the experience of using intelligent systems, users develop a sense of accuracy about the system \cite{Nourani2020TheRO}. When deciding whether to follow the advice of an AI model, individuals can make a better decision if they have acquired a clear understanding and thus an accurate model of the AI system, including its strengths and weaknesses \cite{Bansal2019BeyondAT, Gero2020MentalMO, Holstein2021DesigningFH}. In AI-assisted decision-making, a significant aspect of the user's mental model is recognising errors made by the AI, so that the human can decide when to accept or reject the AI's recommendation \cite{Bansal2020DoesTW}. 
\vspace{-1mm}
\vspace{0.5em}
\begin{minipage}{0.08\textwidth}
  \centering
  \includegraphics[width=0.75\textwidth]{images/Interactions1.pdf}
\end{minipage}%
\begin{minipage}{0.9\textwidth}
 \vspace{1em}
  \subsubsection{Interactions with other Determinants} 
\end{minipage}
\vspace{-1mm}

\textbf{Mental Model, Transparency, and Explainability of AI}. In the context of AI-assisted decision-making, the influence of explanations on mental models is crucial for enhancing comprehension and fostering accurate understanding \cite{Klein2008Macrocognition}. Research has shown that explainable AI has the potential to significantly improve users' understanding of black-box models \cite{lakkaraju2017interpretable}. Explanations characterized by both high soundness and completeness prove to be the most effective in aiding participants' understanding of an intelligent agent's functionality and in improving the accuracy of users' mental models \cite{Kulesza2013Toomuch}. The impact of explanations on mental models can be analyzed through two key processes: maintenance and building. In the mental model maintenance process, individuals tend to uphold or strengthen their existing beliefs. When faced with new information, they interpret or integrate it in a way that aligns with their current understanding. This implies that well-crafted explanations can help users reinforce their existing mental models and ensure they are aligned with the AI system's decision rationale. On the other hand, the mental model building process occurs when individuals undergo substantial restructuring or create entirely new mental models in response to novel or contradictory information \cite{Bauer2023Expl(AI)ned}. 
\\
\\
\textbf{Mental model and Trust in/ Reliance on AI-Systems}. Incomplete or inaccurate mental models can lead to inappropriate reliance and trust \cite{Steyvers2023Three}, resulting in over- or under-reliance \cite{Nourani2021AnchoringBA}. Users' trust might rise the more time they have spent interacting with an AI system, probably due to the fact that they understand the system better \cite{Elkins2013TheSound, Lee2021Brokerbot}. With increased interaction, users might gain insights into the system's capabilities, limitations, and the nuances of its responses, contributing to a more accurate formation of the user's mental model of the intelligent system. In addition to this, research has revealed that expertise, i.e. mental models, and user reliance in intelligent systems are related. In contrast to experienced users, novice users show over-reliance because they do not have the necessary knowledge to identify errors \cite{Nourani2020TheRO}. 

\vspace{-1mm}

\begin{minipage}{0.08\textwidth}
  \centering
  \includegraphics[width=0.70\textwidth]{images/LLM1.pdf}
\end{minipage}%
\begin{minipage}{0.9\textwidth}
 \vspace{1em}
  \subsubsection{Deriving Implications for LLM-assisted Decision-Making} 
\end{minipage}
\textbf{Mental Model of LLMs, Capabilities and Limitations of LLMs}. When making LLM-assisted decisions, building an accurate mental model of the LLM is crucial. Accurate mental models of LLMs probably lead to their correct usage. Inadequate usage of LLMs may stem from incomplete or inaccurate mental models. Hence, knowing the capabilities and limitations of LLMs helps users understand the complexity of these models and how LLMs process information and generate responses. For example, an accurate mental model of prompt engineering enables users to formulate queries and prompts in a manner that aligns with the LLM's capabilities, enhancing the likelihood of obtaining precise and relevant responses. By utilizing appropriate prompting techniques users can structure prompts to extract key information essential for the specific decision-making task. Mental models support users to set expectations about what the LLM is capable to. Moreover, a well-developed mental model allows users to interpret LLM generated outputs effectively and to understand the reasoning behind the LLM's suggestions. Users with an accurate mental model can critically assess LLM-generated content and recognize errors or inconsistencies in LLM-generated content. This critical evaluation is essential for making informed decisions. 
\\
\\
\textbf{Mental Model of Decision Problem and Application Fields of LLMs.} LLMs developed for specific application fields, such as medicine, computational biology, or law, are likely trained on vast amounts of domain-specific data. Therefore, professionals with expertise and experience, i.e., with a more accurate mental model of the decision-problem, can use these LLMs more effectively, as they can precisely define tasks and prompts as well as decision-making requirements within their field. 
\\
\\
\textbf{Mental Model of LLMs, Transparency, and Explainability.} In the realm of LLM-assisted decision-making, transparency assumes a crucial role in influencing users' mental models. Users are more inclined to develop precise representations, i.e., accurate mental models, of how LLMs operate when provided with insights into the underlying mechanisms and considerations. Therefore, explanations have the potential to significantly enhance the understanding of LLMs as black-box models. Specifically, explanations characterized by high soundness and completeness are likely to be the most effective in facilitating participants' comprehension of the functionality of LLMs.
\\
\\
\textbf{Mental Model of and Trust in/ Reliance on LLMs.} Mental models help users establish expectations regarding what LLMs can and cannot do. They assist users in calibrating their trust in the LLM. A user possessing a sophisticated mental model, which includes an understanding of the LLM's underlying mechanisms, strengths, and weaknesses is more likely to trust the LLM appropriately. However, inaccurate user mental models may lead to either over-reliance or under-reliance on LLMs in the decision-making process. Users with inaccurate mental models might overestimate the LLM's capabilities, resulting in over-reliance on LLM suggestions. On the other hand, users with inaccurate mental models could also underestimate the LLMs's abilities, leading to under-reliance.
Mental models are dynamic and can evolve with experience. Decision-makers initially interact with LLMs, observing the responses and suggestions generated by the LLM. This observation includes noting how the LLM interprets queries, the relevance and accuracy of its answers, and proficiency in handling complex or ambiguous requests. From these observations, users form initial perceptions of the LLM's capabilities and limitations. As interactions with the LLM continue, users may refine their mental models based on new experiences. If an LLM consistently offers valuable insights, users might develop a model that regards the LLM as a reliable assistant. Conversely, if the LLM frequently misunderstands queries or provides irrelevant information, users might perceive it as a tool with considerable limitations. This process can be regarded as iterative. Each interaction informs the decision maker's understanding of LLMs, shaping how they approach future interactions and expectations. For example, if a user discovers that an LLM excels in processing data analyses but struggles with creative tasks, they will adjust their usage accordingly. Over time, these mental models help in forming more realistic expectations. Users learn when to rely on the LLM for assistance  and when it's recommended to rely on other sources or their own judgment in the decision-making process.

\vspace{0.5em}
\begin{minipage}{0.08\textwidth}
  \centering
  \includegraphics[width=0.75\textwidth]{images/Scenario.pdf}
\end{minipage}%
\begin{minipage}{0.9\textwidth}
 \vspace{1em}
  \subsubsection{Scenario-based Demonstration} 
\end{minipage}
\textbf{Mental Models, Capabilities, and Limitations of LLMs (Scenario 6).} In the case of Alex, the sales manager, a comprehensive understanding of the capabilities, strengths and limitations of LLMs could have significantly influenced his decision-making process. Recognizing the strengths of LLMs, such as their proficiency in processing extensive volumes of historical sales data and market trends, their adeptness at identifying patterns within data, and their utility in forecasting sales trends based on existing information, is crucial for setting realistic expectations. However, it is equally important to acknowledge the constraints of LLMs. For instance, their reliance on historical data might bias predictions if past trends do not reflect future market conditions. Additionally, LLMs may lack the nuanced understanding of contextual market disruptions that affect data interpretation. Understanding these limitations of LLMs would have prompted Alex to consider external factors that the model might not have accounted for, such as market saturation, competitor actions, changes in consumer behavior, or economic downturns. With an awareness of the LLM's limitations, he would have been more likely to conduct a manual review of the projections, potentially adjusting them based on his own expertise, recent market developments, or insights from other team members.
\\
\\
\textbf{Mental Model of Prompt Engineering (Scenario 1).} In Dr.\ Smith's case, having an accurate mental model of prompt engineering, which involves creating precise and detailed instructions to guide the LLM's responses, could have resulted in more precise and reliable answers. For example, if the prompt provided to the LLM included more specifics about the patient's symptoms, medical history, and relevant contextual information, the LLM could generate a more precise and targeted response. Additionally, the prompt could explicitly direct the LLM to provide a list of potential diagnoses, encompassing both rare and common conditions, along with their probabilities based on the symptoms described. By instructing the LLM to prioritize the likelihood of common ailments first, Dr.\ Smith could consider more probable possibilities before exploring rarer diagnoses. 
\\
\\
\textbf{Mental Model of and Reliance on LLMs (Scenario 4).} In Paula's case, seeking medical advice for vaccination, being aware that LLMs generate responses based on their training data and lack the ability to differentiate between credible and pseudo-scientific information could have resulted in a more well-informed decision. This awareness is particularly important in cases of potential over-reliance on such technology. Paula could use resources like LLMs to gather general information about vaccinations. However, it would be crucial for her to cross-verify this information with reputable medical sources, consult healthcare professionals, and rely on evidence-based research for medical decisions. 

\vspace{-3mm}
\vspace{1,5em}
\subsection{Information Processing}
\label{subsec: Information Processing}
\vspace{-1mm}

\begin{minipage}{0.09\textwidth}
  \centering
  \includegraphics[width=0.7\textwidth]{images/Books.pdf}
\end{minipage}%
\begin{minipage}{0.9\textwidth}
  \subsubsection{Theoretical Background}
\end{minipage}
Thinking is crucial for decision-making as it involves the process of gathering, analyzing, and evaluating information. Consequently, the selection of the final decision among several alternatives is a result of thinking \cite{Pathak2020EffectOE}. Dual-process theories propose two qualitatively distinct processes underlying \texttt{Information Processing} and decision-making \cite{DeNeysDual&SingleModelsofThinking2021, GawronskiDualProcess2021}: \texttt{Intuitive Thinking} and \texttt{Deliberate Thinking}. One process is a fast and intuitive, with high capacity and minimal cognitive effort, while the other is slow and deliberative, with low capacity and high cognitive demand \cite{BelliniDualProcess2022} \cite{EvansIntuition&Reasoning2010}. Judgments and decisions are considered to result from a continuum of processes, ranging from relatively fast and intuitive, to slow and deliberate \cite{ThompsonAnalyicThinking2012}. 

Dual-process theory posits that individuals predominantly think intuitively, often relying on heuristics in decision-making \cite{kahneman2017thinking}. Heuristics can be understood as cognitive shortcuts used consciously or unconsciously to decrease the complexity of decision-making \cite{EysenckcognitvePsychology2020}. Intuitive processing thus can create efficient responses, enabling the decision-maker to make decisions saving a considerable amount of time and cognitive resources. However, this often occurs without analytic processing and can lead to reduced precision \cite{CeschiDimensions_2019, Croskerry2013Deciding}. Many daily decisions can be successfully made using heuristics, and analytical thinking is seldom triggered due to its slower and more resource-consuming nature. While heuristics can be useful, they may lead to cognitive biases, resulting in incorrect and sub-optimal decisions \cite{Tversky1974Judgment}. A cognitive bias is referred as “systematic error in judgment and decision-making common to all human beings which can be due to cognitive limitations, motivational factors, and/or adaptations to natural environments” \cite{Kahneman1982Judgement}. 

Although AI offers numerous advantages, such as its ability to process vast amounts of data, it can evoke human biases. For instance, AI-assisted decision-making tasks are prone to \textit{Anchoring Bias} \cite{Rastogi2020DecidingFA}. The \textit{Anchoring Bias}, also known as \textit{First Impressions}, occurs when people inappropriately adjust their judgement based on an anchor, an initial piece of information. This bias refers to the tendency to modify one's judgment in the direction of the initial information \cite{Tversky1974Judgment}. This is because making adjustments requires effort, and individuals often stop once they reach a reasonably plausible estimate \cite{Tversky1974Judgment}. Research has indicated that the \textit{Anchoring Bias} can negatively impact the overall performance of human-AI-teams when the AI's suggestions are incorrect. However, dedicating more time to AI-assisted decision-making can reduce the influence of the \textit{Anchoring Bias} \cite{Rastogi2020DecidingFA}.

The \textit{Confirmation Bias} is considered as the tendency to seek, interpret, favor, and recall information in a way that confirms one's pre-existing beliefs \cite{nickerson1998confirmation} and to discount contradictory information \cite{Moynihan2010Cognitive}. It leads individuals to selectively focus on information that fits with their existing beliefs, disregarding alternative perspectives. This bias impedes a comprehensive review of all available information, often resulting in inappropriate decision-making \cite{Huand2012Understanding}.

\begin{minipage}{0.08\textwidth}
  \centering
  \includegraphics[width=0.75\textwidth]{images/Interactions1.pdf}
\end{minipage}%
\begin{minipage}{0.9\textwidth}
 \vspace{1em}
  \subsubsection{Interactions with other Determinants} 
\end{minipage}
\textbf{Information Processing and Mental Models.} Mental models serve as guides for information search, aiding individuals in assessing the relevance of information, understanding it, and deciding whether to discard irrelevant or unhelpful information \cite{festinger1962theory}. Simultaneously, mental models can constrain information acquisition, search, filtering, coding, storage, and retrieval. Consequently, valuable information might be excluded and not considered in subsequent decision-making. Therefore, mental models can lead to cognitive biases, prompting individuals to disregard information perceived as insignificant, even if it might be crucial \cite{Huand2012Understanding}. Decision-makers tend to actively seek evidence that confirms their beliefs while neglecting or discounting contradictory information \cite{Arnott2006Cognitive}, known as \textit{Confirmation Bias} \cite{Wason1960OnTF}. The \textit{Confirmation Bias} can impede decision-making as individuals may confine themselves to favored hypotheses and neglect alternative possibilities \cite{nickerson1998confirmation}.
\\
\\
\textbf{Information Processing and Trust in/ Reliance on AI.} Individuals frequently rely on intuitive thinking, employing heuristics and shortcuts in decision-making \cite{kahneman2017thinking}. Interrupting heuristic thinking and, consequently, engaging in analytical reasoning can help decrease over-reliance on AI-generated explanations in the context of AI-assisted decision-making \cite{Buccinca2021ToTO}. Cognitive forcing strategies, namely interventions designed to interrupt heuristic thinking and engage users in effortful, analytic thinking \cite{Lambe2016Dual-process}, can decrease over-reliance on AI in AI-assisted decision-making \cite{Buccinca2021ToTO}.  However, there are also situations where humans rarely rely on the suggestions generated by the AI system \cite{Vasconcelos2022ExplanationsCR}. For instance, individuals seldom consider a suggested automatic reply for an important email to their supervisor \cite{Hancock2020AI-Mediated}. Drawing on cost-benefit frameworks \cite{Navon1977OnTE, Vasconcelos2022ExplanationsCR} over-reliance on AI can be considered the result of a strategic decision. Accordingly, a person is less likely to over-rely on AI suggestions if the benefit of obtaining a correct answer is high. Conversely, a person might be less inclined to over-rely on AI explanations if the costs associated with determining the correct response are low. 

\begin{minipage}{0.08\textwidth}
  \centering
  \includegraphics[width=0.7\textwidth]{images/LLM1.pdf}
\end{minipage}%
\begin{minipage}{0.9\textwidth}
 \vspace{1em}
  \subsubsection{Deriving Implications for LLM-assisted Decision-Making}
\end{minipage}
In the context of LLM-assisted decision-making, the \textit{Anchoring Bias} can significantly influence users' perceptions and choices. The \textit{Anchoring Bias} occurs when individuals rely too heavily on the first piece of information when making decisions. Therefore, in LLM-assisted decision-making, the model's first output could act as an anchor for users. Users might not adjust the LLM's recommendation sufficiently from the initial suggestion. Consequently, they might accept the initial suggestion without critically evaluating its validity or exploring alternative possibilities. This could lead to sub-optimal decisions, as users may not explore the full range of possibilities or adequately consider all relevant information.

\textbf{Information Processing and Reliance on LLMs.} Information processing is likely to play a crucial role in influencing reliance on LLMs in the realm of LLM-assisted decision-making.  Given the common tendency of individuals to resort to intuitive thinking in decision-making, there is a potential risk of over-relying on LLM-generated explanations or suggestions. Conversely, actively participating in analytical reasoning can emerge as a mitigating factor against such over-reliance, suggesting that the way individuals process information directly impacts their reliance on LLMs. Cognitive forcing strategies could encourage individuals to engage in more deliberate, analytical thinking, potentially preventing over-reliance on LLM generated recommendations. 
\\
\\
\textbf{Information Processing and Mental Models.} When utilizing LLMs in decision-making, it's essential to understand how mental models influence information search, aiding in assessing relevance and guiding decisions on discarding irrelevant data. Nevertheless, it is important to acknowledge that mental models can impose constraints on information acquisition from LLMs, potentially leading to the exclusion of valuable data and resulting in cognitive biases. The \textit{Confirmation Bias}, characterized by the tendency to selectively favor information that aligns with pre-existing beliefs, i.e. users' mental models, might significantly impact LLM-assisted decision-making. When users interact with LLMs, they may pay more attention to information from the model that confirms their existing beliefs or expectations. Users might be more inclined to accept information that aligns with their current thinking or beliefs, while ignoring or downplaying conflicting perspectives. Such selective focus hinders a comprehensive understanding of the decision-making situation, potentially leading to sub-optimal outcomes.

\begin{minipage}{0.08\textwidth}
  \centering
  \includegraphics[width=0.7\textwidth]{images/Scenario.pdf}
\end{minipage}%
\begin{minipage}{0.9\textwidth}
 \vspace{1em}
  \subsubsection{Scenario-based Demonstration} 
\end{minipage}
\textbf{Anchoring Bias in LLM-assisted Decision-Making (Scenario 5).} Jenna, the marketing manager, approached an LLM to optimize her company's digital advertising strategy. In Jenna’s case the \textit{Anchoring Bias} could have influenced her decision, as the LLM's initial recommendation to invest in the new social media platform could have served as the anchor for Jenna's decision-making process. The anchor could have restricted Jenna's consideration of alternative platforms or strategies. Even if other platforms may offer better value or alignment with the company’s goals, the influence of the initial suggestion may limit exploration. Moreover, Jenna might have been more inclined to seek information that supports the chosen social media platform, potentially overlooking data that suggests a different allocation could be more effective.
\\
\\
\textbf{Confirmation Bias in LLM-assisted Decision-making (Scenario 2).} David, already concerned about his health and seeking answers in an LLM-powered medical advice forum, may have paid more attention to information that aligned with his fears or suspicions. If the LLM suggests a rare illness, he might be more inclined to focus on and remember that information, potentially overlooking more likely or common explanations for his symptoms. David might be more likely to seek out additional information that supports the suggestion made by the LLM. This could involve further online searches, consulting additional medical sources that also suggest the rare illness, or even discounting information that contradicts the initial suggestion.

\vspace{-3mm}
\vspace{1,5em}
\subsection{Emotions and Mood}
\label{subsec: Emotions and Mood}
\vspace{-1mm}

\begin{minipage}{0.09\textwidth}
  \centering
  \includegraphics[width=0.7\textwidth]{images/Books.pdf}
\end{minipage}%
\begin{minipage}{0.9\textwidth}
  \subsubsection{Theoretical Background}
\end{minipage}

Affect encompasses both \texttt{Emotion and Mood}, each distinguished by their respective object and temporal limitations. An emotion usually arises in response to a certain eliciting stimulus or event, exhibiting intensity but limited duration. On the other hand, mood is generally not tied to a specific stimulus, characterized by lower intensity and longer duration \cite{schnall2011affect}.

Emotions play a major role in influencing decision making. Key findings from research on emotions and decision-making include \cite{Lerner2015Emotion}: (1) Emotions exert a potent, predictable, sometimes harmful, and sometimes beneficial impact on decision-making, affecting different types of decisions. (2) Emotions can generate effects that are undesirable and subconscious. (3) Initially, emotions are often triggered swiftly, leading to rapid actions and (4) once activated, certain emotions, such as sadness, can promote more systematic thinking.

\begin{minipage}{0.08\textwidth}
  \centering
  \includegraphics[width=0.75\textwidth]{images/Interactions1.pdf}
\end{minipage}%
\begin{minipage}{0.9\textwidth}
 \vspace{1em}
  \subsubsection{Interactions with other Determinants} 
\end{minipage}
\textbf{Emotions and Mood, and Information Processing.} The \textit{Affect-as-Information Hypothesis} suggests that emotions and mood act as sources of information \cite{Clore1992Cognitive}. The depth of processing is influenced by the presence of positive and negative emotions, which activate either heuristic or systematic information processing in decision-making \cite{Pathak2020EffectOE}. Consequently, individuals in a positive mood perceive their environment as benevolent, leading them to process information in a global and heuristic manner. Conversely, those in a negative mood view their environment as problematic, prompting them to process information analytically and diagnostically. The \textit{Affect-as-Information Hypothesis} is often relied upon when the target is affective, the information is complex, or time constraints exist \cite{Clore1994Affective}. Similarly, the \textit{Affect Infusion Model} postulates that individuals process information corresponding with their mood state. This means that people in a negative mood tend to process information more accurately, analytically, and in more detail, while those in a positive mood engage in more simplified and heuristic processing \cite{Forgas1995Mood}. Research has also indicated that individuals in a negative mood are inclined to process information more systematically, whereas those in a positive mood are more likely to utilize heuristic processing \cite{Mohanty2014Decision}.
\\
\\
\textbf{Emotions and Trust in AI.} Human emotions can influence trust in AI and the use of AI. Emotional trust, such as feelings of safety, well-being, and satisfaction, can be distinguished from cognitive trust, i.e. people's rational expectation that AI can perform well \cite{Komiak2006TheEffects}. Incidental emotions, which are emotions not directly related to the task at hand, can significantly impact trust in unrelated settings. Emotions characterized by positive valence, such as happiness and gratitude, may enhance trust. Conversely, emotions with negative valence, such as anxiety or anger, might reduce trust \cite{Dunn2003Feeling}.

\begin{minipage}{0.08\textwidth}
  \centering
  \includegraphics[width=0.7\textwidth]{images/LLM1.pdf}
\end{minipage}%
\begin{minipage}{0.9\textwidth}
 \vspace{1em}
  \subsubsection{Deriving Implications for LLM-assisted Decision-Making} 
\end{minipage}
Drawing on the \textit{Affect-as-Information Hypothesis} \cite{Clore1992Cognitive} and on the \textit{Affect Infusion Model} \cite{Forgas1995Mood} emotional states might influence how decision-makers process information provided by LLMs. Hence, if a decision-maker is in a positive mood or experiencey positive feelings, they might be more inclined to process the LLM's suggestions more intuitively and heuristically. This might lead them to accept LLM's recommendations more uncritically. Conversely, if a decision-maker is in a negative emotional state, they might engage in more deliberate and critical information processing, leading to a deeper evaluation of LLM generated content.

The influence of human emotions on trust and the use of AI, particularly in the context of LLM-assisted decision-making, is a multifaceted issue. Emotional trust can directly influence the degree of reliance placed on LLMs. When users feel emotionally secure and satisfied with an LLM, they are more likely to trust its recommendations and use it as a decision-making aid. For instance, if a user feels that an LLM consistently understands and responds to their queries effectively, they may develop a sense of safety and well-being around its use, leading to increased trust and reliance on the system for decision support. Conversely, negative emotions, like anger or frustration, can diminish trust in LLMs. If a user encounters repeated errors or feels that the system does not understand their needs, this can lead to frustration or anger, emotions with negative valence. Such emotions might cause the user to question the LLM's capabilities, thereby reducing their trust and willingness to use the LLM for decision-making.

Incidental emotions can also significantly impact trust in LLMs. For example, a user experiencing general happiness or gratitude  may be more inclined to trust and use an LLM, even in unrelated decision-making scenarios. This positive emotional state can lead to a more optimistic and accepting attitude towards the LLM's suggestions. On the other hand, if a user is experiencing incidental negative emotions, such as anxiety or sadness from unrelated life events, they might project these feelings onto their interactions with the LLM. This could manifest as increased skepticism or reduced trust in the LLM's capabilities, thereby affecting their reliance on the LLM for decision-making.

\vspace{-2mm}

\begin{minipage}{0.08\textwidth}
  \centering
  \includegraphics[width=0.75\textwidth]{images/Scenario.pdf}
\end{minipage}%
\begin{minipage}{0.9\textwidth}
 \vspace{1em}
  \subsubsection{Scenario-based Demonstration} 
\end{minipage}
\textbf{Positive Emotions and Information Processing (Scenario 6).} In Alex's situation, positive affect can impact how he interprets and assesses the information. When in a positive mood, he may employ simplified and heuristic processing, increasing the likelihood of accepting the LLM-generated sales projections without thorough examination or critical evaluation. Moreover, positive affect can contribute to decision biases, such as \textit{Confirmation Bias}, where individuals tend to favor information that confirms their existing beliefs. In this scenario, Alex's positive emotions might cause him to overlook or downplay contradictory information, reinforcing his initial trust in the LLM-generated projections and inhibiting critical evaluation.
\\
\\
\textbf{Negative Emotions and Information Processing (Scenario 3).} In Anna's case, who is looking for a weight loss adivce, negative emotions could lead to a more critical and cautious approach to processing recommendations from LLMs. Experiencing negative emotions might motivate Anna to seek additional information to validate or refute the LLM's advice. She might start looking for more balanced, scientifically-backed dietary advice, or consult healthcare professionals for a more personalized assessment. While positive emotions might have initially led her to pay attention to the potential benefits of the diet, negative emotions might redirect her attention to its risks and drawbacks. This shift can lead to a more critical evaluation of the information provided by the LLM.
\\
\\
\textbf{Positive Emotions and Trust in LLMs (Scenario 5).} In the scenario where Jenna, a marketing manager at a tech startup, consults an LLM for assistance in optimizing the company's digital advertising strategy, her initial positive feelings can significantly influence her trust in the LLM-assisted decision-making process. Her initial positive emotions could predispose her to be more receptive to the LLM's suggestions and lead her to view the LLM's recommendation to invest in a new social media platform as a creative and forward-thinking solution. Positive feelings can enhance Jenna's trust in the LLM's capabilities. Believing in the LLM's ability to analyze market trends and identify promising opportunities, she might be more inclined to trust its recommendation about the new social media platform.
\\
\\
\textbf{Negative Emotions and Trust in LLMs (Scenario 5).} In the scenario where Paula, an expectant mother, consults an LLM-powered online forum for medical advice regarding her newborn's vaccination, and the forum provides misguided advice based on pseudo-scientific information, her initial negative feelings could significantly impact her trust in LLM-assisted decision-making. If Paula initially approaches the LLM-powered forum with negative emotions, such as concern, these adverse feelings can predispose her to distrust the information provided. 
\vspace{-3mm}

\vspace{1,5em}
\subsection{Metacognitions}
\vspace{-1mm}

\label{subsec: Metacognitions}
\begin{minipage}{0.08\textwidth}
  \centering
  \includegraphics[width=0.7\textwidth]{images/Books.pdf}
\end{minipage}%
\begin{minipage}{0.9\textwidth}
  \subsubsection{Theoretical Background}
\end{minipage}
\texttt{Metacognitions} are regarded as the processes that monitor ongoing thought processes and control the allocation of mental resources \cite{AckermanMetaReasoning2017}. Metacognitions refer to "thinking about one's own thinking - the capability to detach oneself from one's own thinking, observe it objectively, and identify opportunities to apply strategic thinking interventions" \cite{Croskerry2003Cognitive}.

Meta-reasoning focuses on "the processes that monitor and control reasoning, problem solving, and decision making" \cite[p.~275]{thompsonMeta-reasoning2016}. These metacognitive processes act as the "top manager" of cognitive functions, responsible for regulating functions, such as setting goals, selecting among reasoning strategies, and making decisions \cite{FiedlerMetacognition2019}. \texttt{Monitoring} is defined as the subjective assessment of the quality of performance in a cognitive task. In contrast, metacognitive \texttt{Control} involves initiating, terminating, or modifying the effort allocated to a cognitive task (\cite{AckermanMetaReasoning2017}. Monitoring can manifest in various forms of judgments that individuals spontaneously exhibit before, during, or after cognitive processing. These judgments might include assessing whether a task is feasible, evaluating progress, or estimating the likelihood of success of a particular choice \cite{AchermanHeuristicCues2019}. Monitoring processes run in the background, reflecting states of certainty or uncertainty. Such metacognitive processes exert a control over the degree of mental effort applied. If a person is confident in an answer or a choice, they will likely choose it. Conversely, if a person feels uncertain, they may seek additional information or consider a different approach \cite{AckermanMetaReasoning2017}. 

Judgments and decisions are often accompanied by a sense or \texttt{Feeling of Rightness} or a \texttt{Feeling of Error}. This process is highly effective, as continuously verifying the accuracy of judgments and decisions would be extremely resource-intensive. Identifying something as true or false requires minimal effort, as the corresponding feeling is automatically elicited by the decision \cite{NadurakDualProcessTheory2023}.

\begin{minipage}{0.08\textwidth}
  \centering
  \includegraphics[width=0.75\textwidth]{images/Interactions1.pdf}
\end{minipage}%
\begin{minipage}{0.9\textwidth}
 \vspace{1em}
  \subsubsection{Interactions with other Determinants} 
\end{minipage}
\textbf{Metacognitions, Information Processing, and Reliance on LLMs.} Metacognitions can trigger adverse emotions, such as a feeling of uncertainty or error, when dealing with uncertain or potentially incorrect information that requires revision \cite{ArangoScaffoldedMemory2013}. The \textit{Feeling of Rightness (FOR)} can impact subsequent behaviors and predict the likelihood of changing answers later \cite{Wang2019Fluency}. When the FOR is weak, it triggers analytical problem-solving and extended deliberation. In contrast, a strong FOR indicates that further reflection and reconsideration of the answer are unnecessary \cite{AckermanMetaReasoning2017}, likely resulting in over-reliance on AI. The \textit{Feeling-of-Error (FOE)} can signal to individuals a possible failure in their mental processes. The subjective experience that something has gone wrong provides the basis for making subsequent corrections and improving reasoning, and encourages specific behaviors, such as modifying a strategy or reviewing the outcome of a mental action \cite{FerndandezOppsScratsch2016}.

\begin{minipage}{0.08\textwidth}
  \centering
  \includegraphics[width=0.7\textwidth]{images/LLM1.pdf}
\end{minipage}%
\begin{minipage}{0.9\textwidth}
 \vspace{1em}
  \subsubsection{Deriving Implications for LLM-assisted Decision-Making} 
\end{minipage}
\textbf{Metacognitions, Information Processing, and Reliance on LLM.} A strong FOR might lead decision-makers to conclude that further reflection or reconsideration of the LLM's generated answer is redundant. This situation poses a risk of over-reliance, where users might overlook potential errors, biases, or incomplete information in the LLM's responses. Conversely, when individuals experience a strong FOE while interacting with an LLM, it might heighten their awareness that something might be wrong with the information or reasoning provided by the LLM. Thus, the FOE can encourage users to question the LLM's responses more critically and decision-makers are more likely to be prompted to consider the possibility of inaccuracies or misinterpretations in the LLM's answers. As a FOE can foster a more critical engagement with LLMs generated output, it might prevent over-reliance.
In Figure \ref{fig:psych-determinants}, potential metacognitions are depicted.

\begin{minipage}{0.08\textwidth}
  \centering
  \includegraphics[width=0.75\textwidth]{images/Scenario.pdf}
\end{minipage}%
\begin{minipage}{0.9\textwidth}
 \vspace{1em}
  \subsubsection{Scenario-based Demonstration} 
\end{minipage}
\textbf{Feeling of Rightness, Information Processing, and Reliance on LLMs (Scenario 2).} In this scenario, David, a concerned patient experiencing unexplained symptoms, turns to an LLM-powered medical advice forum. Upon receiving a suggestion of a rare illness from the LLM, if David experiences a strong FOR, he may accept the LLM's suggestion without engaging in further analytical information processing, such as seeking additional medical consultation. Consequently, a strong FOR may result in an over-reliance on the LLM's suggestion, preventing Alex from exploring other potential explanations for his symptoms.

\textbf{Feeling of Error, Information Processing, and Reliance on LLMs (Scenario 6).} In the scenario where Alex, a sales manager, uses an LLM to generate sales projections for the upcoming quarter, experiencing a FOE while reviewing the sales forecasts provided by the LLM, might encourage Alex to adopt a more analytical approach to decision-making, and to critically evaluate the information. Triggered by the FOE, Alex might seek additional verification of the LLM's forecasts. The FOE can also serve as a safeguard against over-reliance on the LLM's capabilities, reminding Alex that LLMs are not infallible.
\vspace{-3mm}

\vspace{1,5em}
\subsection{Decision-Making Styles}
\vspace{-1mm}

\label{subsec: Decision-making styles}
\begin{minipage}{0.08\textwidth}
  \centering
  \includegraphics[width=0.7\textwidth]{images/Books.pdf}
\end{minipage}%
\begin{minipage}{0.9\textwidth}
  \subsubsection{Theoretical Background}
\end{minipage}
Three distinct \texttt{Decision-Making Styles} can be identified: \texttt{Maximizers}, \texttt{Satisficers}, and \texttt{Minimizers}. Maximizers perform a detailed comparison of all available options, aiming for the optimal decision outcome. In contrast, satisficers rely on a limited set of criteria to make a decision and are satisfied with options that meet these criteria. Minimizers, on the other hand, conduct a less accurate search than satisficers and terminate it sooner. They use a single criterion to evaluate the options \cite{MiscuraDecisionMakingTendency2015}, focusing on minimizing the resources invested in decision-making to achieve the minimum decision result \cite{Rogge2022Exploring}. Research has revealed that decision styles remain relatively consistent across various decision domains, with individuals tendency to consistently maximize or satisfice \cite{MoyonoMaximizers2021}. Compared to satisficers, maximizers are inclined to invest greater effort in searching for alternatives, striving to achieve the best possible outcomes \cite{dar2009maximization, weaver2015role}. 

\begin{minipage}{0.08\textwidth}
  \centering
  \includegraphics[width=0.75\textwidth]{images/Interactions1.pdf}
\end{minipage}%
\begin{minipage}{0.9\textwidth}
 \vspace{1em}
  \subsubsection{Interactions with other Determinants} 
\end{minipage}
\textbf{Decision-Making Styles and Information Processing}. Minimizers, who prioritize minimizing resources in decision-making, are likely to employ fast information processing. Consequently, they tend to rely on fast, intuitive judgments and heuristics for swift decision-making. Satisficers, who seek options that fulfill particular criteria, might use a combination of fast and deliberate information processing, probably influenced by the complexity of the decision. For simpler decisions, satisficers might use fast processing, relying on heuristics and readily available information to quickly identify satisfactory options. In contrast, for more complex decisions, satisficers may shift to slow processing, taking the time to gather and analyze detailed information to ensure the chosen option meets their criteria adequately. Maximizers, who strive for the the best possible outcome, are inclined to predominantly engage in slow information processing. They are likely to invest considerable time and effort in thouroughly researching, evaluating, and considering all available information to ensure they make the most optimal decision.

\begin{minipage}{0.08\textwidth}
  \centering
  \includegraphics[width=0.7\textwidth]{images/LLM1.pdf}
\end{minipage}%
\begin{minipage}{0.9\textwidth}
 \vspace{1em}
  \subsubsection{Deriving Implications for LLM-assisted Decision-Making}
\end{minipage}
\textbf{Decision-Making Styles and Information Processing.} The decision-making styles of minimizers, satisficers, and maximizers are likely to influence their interactions with and utilization of LLMs in their decision-making processes. Minimizers, who prefer quick and efficient decision-making, might readily accept the first LLM suggestion that meets their requirements and do not further analyze other LLM generated recommendations. The approach of satisficers may vary depending on the complexity of the decision. For simple decisions, they might use LLMs similarly to minimizers, seeking quick answers. However, for more complex decisions, they would expect the LLM to provide more detailed and carefully analyzed information. Since satisficers search for options that meet specific criteria, they would utilize LLMs to filter and present options that align with these criteria. Maximizers, who aim for the best possible outcome, involve themselves in thorough research and evaluation. Consequently, they would use LLMs for in-depth information gathering. This likely leads to extended interactions with the LLM, as they probe various aspects of a decision, compare options, and weigh the pros and cons.
\\
\\
\textbf{Decision-Making Styles and Over-reliance in LLM-assisted decision-making.} As minimizers aim to minimize their resources and effort spent on decision-making, they might tend to overrely on LLMs for fast decisions. This over-reliance may occur because LLMs offer data-driven, and seemingly optimal decisions, aligning with the goal of minimizing the decision-making process. Likewise, satisficers might be more prone to over-relying on LLMs as they seek decisions that meet specific criteria. This over-reliance could stem from their willingness to settle for options that meet these criteria, making them more inclined to accept AI-generated suggestions without extensive evaluation. Unlike satisficers, maximizers seek the optimal decision across a range of options. Therefore, maximizers may use LLM generated suggestions as a tool to explore a broader array of possibilities and information efficiently. Hence, they might be less likely to over-rely on LLMs because their pursuit of the best possible outcome drives them to critically evaluate AI suggestions and consider multiple sources of information.

\begin{minipage}{0.08\textwidth}
  \centering
  \includegraphics[width=0.75\textwidth]{images/Scenario.pdf}
\end{minipage}%
\begin{minipage}{0.9\textwidth}
 \vspace{1em}
  \subsubsection{Scenario-based Demonstration}
\end{minipage}
\textbf{Decision-Making Styles and Information Processing (Scenario 3).} Anna sought weight loss advice on a website providing recommendations from an LLM that suggested following an extreme diet, eliminating the consumption of almost all carbohydrates. Given Anna's preference for quick and efficient decisions, as a minimizer, she might readily accept the LLM's suggestion of the extreme diet. Her decision-making process tends to be intuitive. As a minimizer, Anna is inclined to trust the LLM's suggestion without engaging in detailed analysis or seeking additional information. If Anna adopted a satisficing approach, she would use the LLM to filter options aligned with specific criteria, emphasizing the need for more nuanced information. These criteria might include considerations such as dietary preferences and potential health risks associated with extreme diets. On the other hand, if she exhibited maximizing tendencies, Anna would aspire to achieve the best possible outcome in her weight loss journey. In such a scenario, she would conduct thorough research and evaluation using the LLM. Anna would engage into in-depth information gathering, probing various aspects of the decision to follow the extreme diet. In this case, Anna might search information, including nutritional impact, scientific evidence, long-term effects, and alternative approaches. 
\\
\\

\vspace{-3mm}
\section{Decision-specific Determinants of LLM-assisted Decision Making}
\label{sec:ds-determinants}
\vspace{-1mm}

\begin{wrapfigure}{r}{0.5\textwidth}
\centering
    \includegraphics[width=0.48\textwidth]{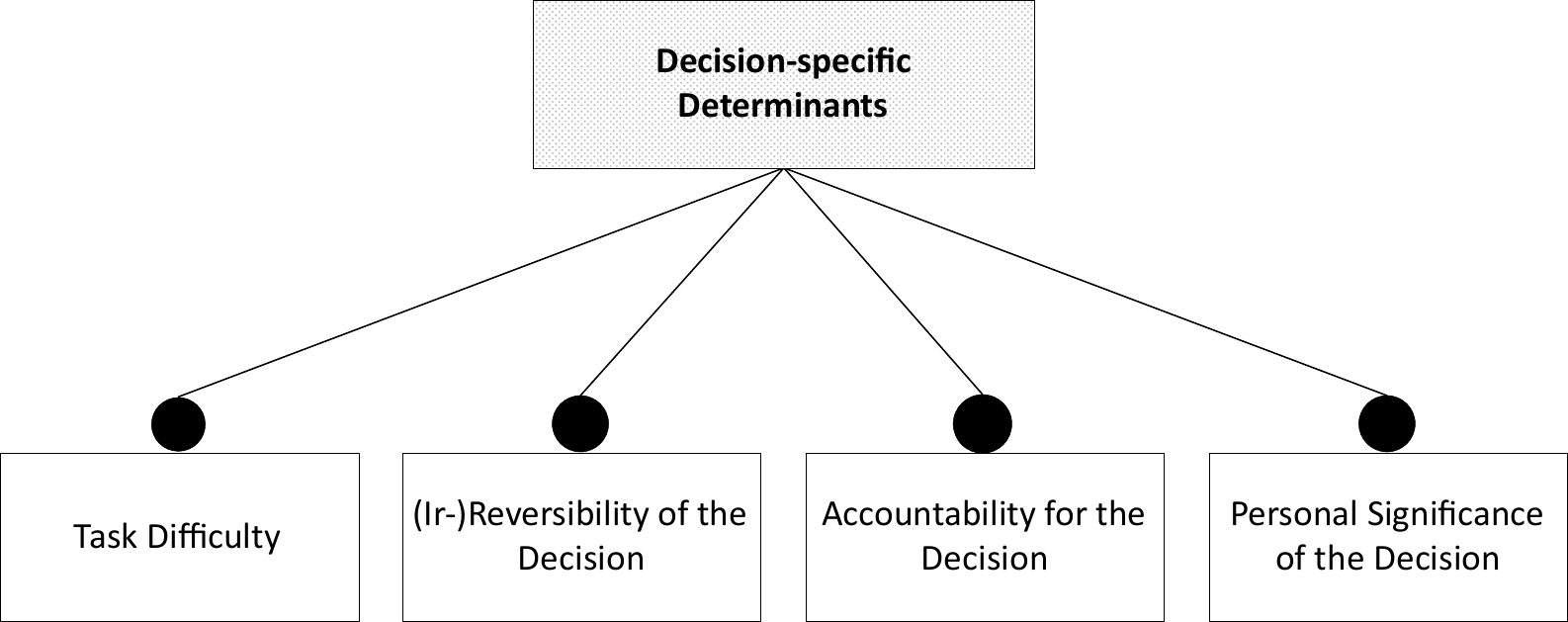}
    \caption{Decision-specific determinants of LLM-assisted decision-making.}
  \label{fig:Decision-specific Determinants of LLM-supported Decision-Making}
\end{wrapfigure}

In this section, we discuss decision-specific determinants. As illustrated in Figure \ref{fig:Decision-specific Determinants of LLM-supported Decision-Making}, \texttt{Task Difficulty}, \texttt{(Ir-)reversibility of the Decision}, \texttt{Accountability for the Decision}, and \texttt{Personal Significance of the Decision} mandatorily influence LLM-assisted decisions.

\begin{minipage}{0.08\textwidth}
  \centering
  \includegraphics[width=0.7\textwidth]{images/Books.pdf}
\end{minipage}%
\begin{minipage}{0.9\textwidth}
   \vspace{1em}
  \subsection{Theoretical Background} 
\end{minipage}

The type of decision problem (e.g., complexity or unfamiliarity) and the characteristics of the decision environment (e.g., the irreversibility of the decision, accountability, subjective significance) impact the choice of strategy in decision-making situations, varying from non-analytical, heuristic to highly analytical strategies \cite{Beach1978AContingencyModel}. Complex situations are characterized by the presence of numerous elements or variables, necessitating the processing of vast amounts of information that surpasses the cognitive abilities of even the most intelligent human decision-makers \cite{Koufteros2005Internal}.

\begin{minipage}{0.08\textwidth}
  \centering
  \includegraphics[width=0.75\textwidth]{images/Interactions1.pdf}
\end{minipage}%
\begin{minipage}{0.9\textwidth}
 \vspace{1em}
  \subsection{Interactions with other Determinants} 
\end{minipage}

\textbf{Task Difficulty and Reliance on Decision-aids and DSSs.} While objective task complexity does not seem to affect reliance on decision-aids, subjective task complexity, also referred to as task difficulty, may have an impact \cite{Parkes2017TheEO}. Research has indicated that as tasks become more difficult, individuals tend to over-rely on decision aids, such as advice from algorithms \cite{Bogert2021HumansRM} or decision-support systems \cite{Parkes2017TheEO}. However, other research did not observe a significant effect of task difficulty in AI-assisted decision-making. One explanation for this contradicting finding is that previous research has revealed that the higher the user's expertise, the smaller the influence of task difficulty on user-reliance \cite{Cao2022UnderstandingUR}.
\\
\\
\textbf{Significance of, Reversibility of, and Accountability for a Decision and Information Processing.} The decision strategy becomes more analytical and involves a greater investment of time and effort when (1) the decision is of great importance, (2) the decision cannot be reversed and (3) the decision-maker is accountable for his or her actions, compared to when the opposite conditions apply \cite{MCALLISTER1979}. Hence, the more people feel accountable for a task's outcome, the more they tend to rely on the computerized decision support \cite{VanDongen2013AFranework}. Moreover, an irreversible decision might trigger more analytical information processing than a reversible choice \cite{MCALLISTER1979}]. In situations where the outcomes of a decision cannot be easily reversed, decision-makers are inclined to anticipate regret \cite{Zeelenberg1999AnticipatedRE}, which probably leads to a negative emotional state and, hence, causes individuals to engage in more deliberate information processing \cite{Mohanty2014Decision}. Likewise, information is more likely to be processed analytically in the decision-making process if the decision is subjectively significant for the person \cite{MCALLISTER1979}. 

\begin{minipage}{0.08\textwidth}
  \centering
  \includegraphics[width=0.7\textwidth]{images/LLM1.pdf}
  \label{fig:dein_bild}
\end{minipage}%
\begin{minipage}{0.9\textwidth}
 \vspace{1em}
\subsection{Deriving Implications for LLM-assisted Decision-Making} 
\end{minipage}

\textbf{Task Difficulty, Reliance on LLMs and Mental Model of LLMs.} People may increasingly rely on LLMs for assistance when confronted with difficult tasks. The difficulty of a task might overwhelm an individual's cognitive capabilities, leading them to seek support from LLMs. If a task is perceived as difficult, individuals will tend to over-rely on LLM recommendations in a decision-making process. However, reliance on LLMs appears to be influenced not just by the difficulty of the task, but also by the expertise, i.e., the mental model, of the decision-maker. In areas where specialized knowledge is not essential, the perceived complexity of the task may not lead to increased reliance on LLMs. Conversely, in complex tasks where the user lacks expertise, there tends to be an over-reliance on LLMs. However, as expertise grows, reliance on LLMs may decrease, even with complex tasks, likely because the individual's ability to process and analyze information independently improves.
\\
\\
\textbf{Perceived Irreversibility of, Significance of, and Accountability for a Decision, Information Processing, and Reliance on LLMs.} The perceived irreversibility of a decision and the accountability associated with it may trigger analytical information processing, potentially reducing over-reliance on LLM suggestions in the decision-making process. Since decision-makers tend to engage in more deliberate information processing when making personally significant decisions, over-reliance on LLM recommendations might be mitigated as a result.

\begin{minipage}{0.08\textwidth}
  \centering
  \includegraphics[width=0.75\textwidth]{images/Scenario.pdf}
\end{minipage}%
\begin{minipage}{0.9\textwidth}
 \vspace{1em}
  \subsection{Scenario-based Demonstration} 
\end{minipage}

\textbf{Irreversibility of and Accountability for the decision, and Information Processing (Scenario 6).} In the scenario of Alex, a sales manager using an LLM to generate sales projections, the perceived irreversibility of a decision can significantly impact how Alex approaches the LLM's suggestions. Understanding that incorrect projections could lead to irreversible consequences, such as overstocking, understocking, or misallocation of resources, Alex is likely to engage in more analytical processing. This means that instead of accepting the LLM-generated sales projections at face value, Alex would scrutinize them more closely. By prompting the LLM to explain its reasoning, Alex could critically assess the validity and reliability of the projections. Knowing that Alex will be held accountable for the outcomes based on these projections, he is incentivized to explore in more detail the LLM's analysis. This involves not just understanding how the LLM arrived at its conclusions but also evaluating whether the LLM's analysis aligns with his knowledge and expectations. For instance, if the LLM's projections are overly optimistic or pessimistic compared to Alex's understanding of market trends and historical data, he might question and re-evaluate the inputs or the model's reasoning.

\section{Dependency Framework of Determinants of LLM-assisted Decision-making}
\label{sec:framework}
\begin{figure*}[htb]
	\centering
	\includegraphics[width=0.8\textwidth]{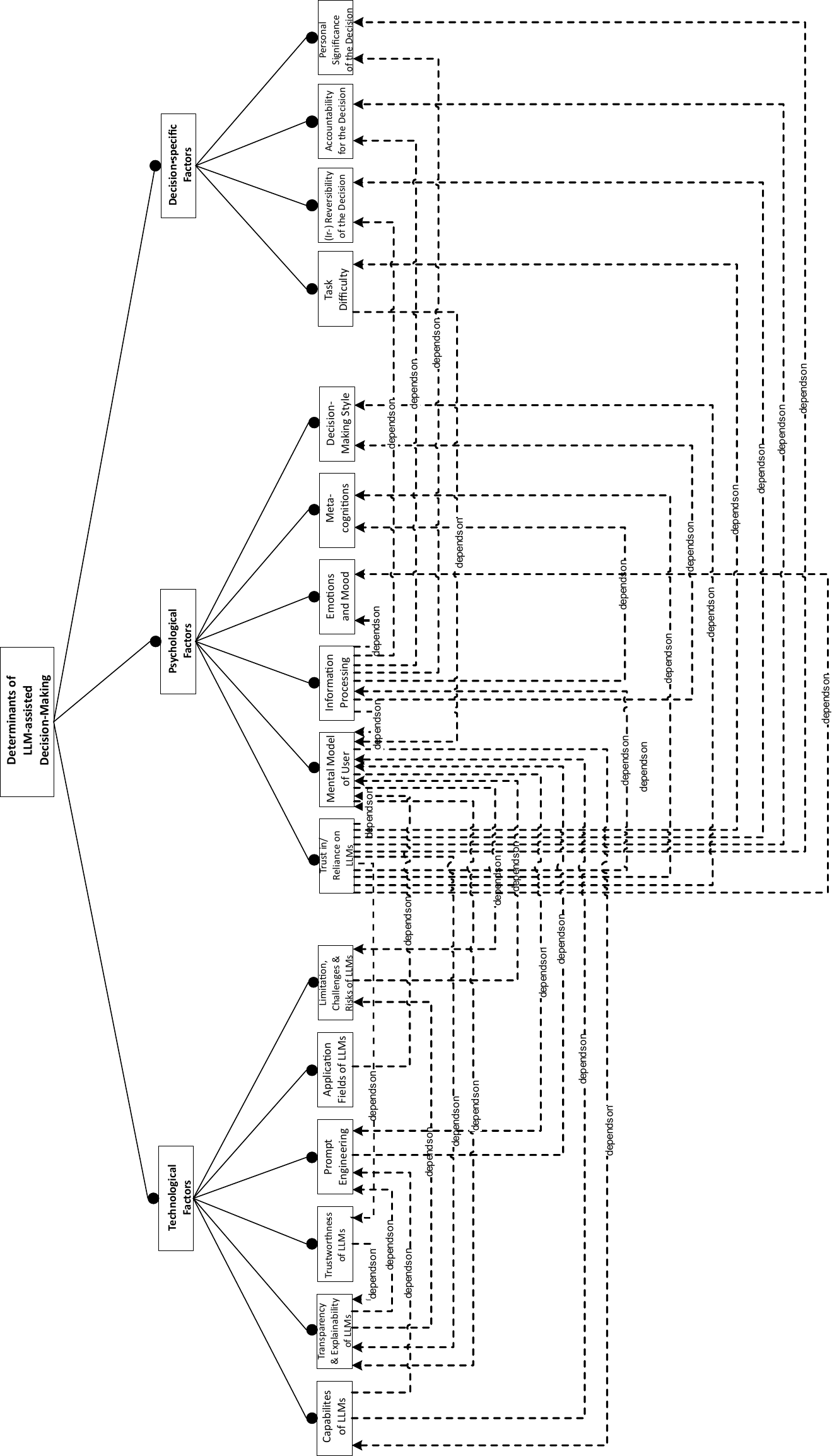}
	\caption{Feature diagram illustrating a structural overview of the identified technological, psychological, and decision-specific determinants of LLM-assisted decision-making as well as of the interdependencies between these determinants.}
	\label{fig:Determinants LLM-supported Decision Making}
\end{figure*}

\begin{figure*}[htb]
	\centering
	\includegraphics[width=1.0\textwidth]{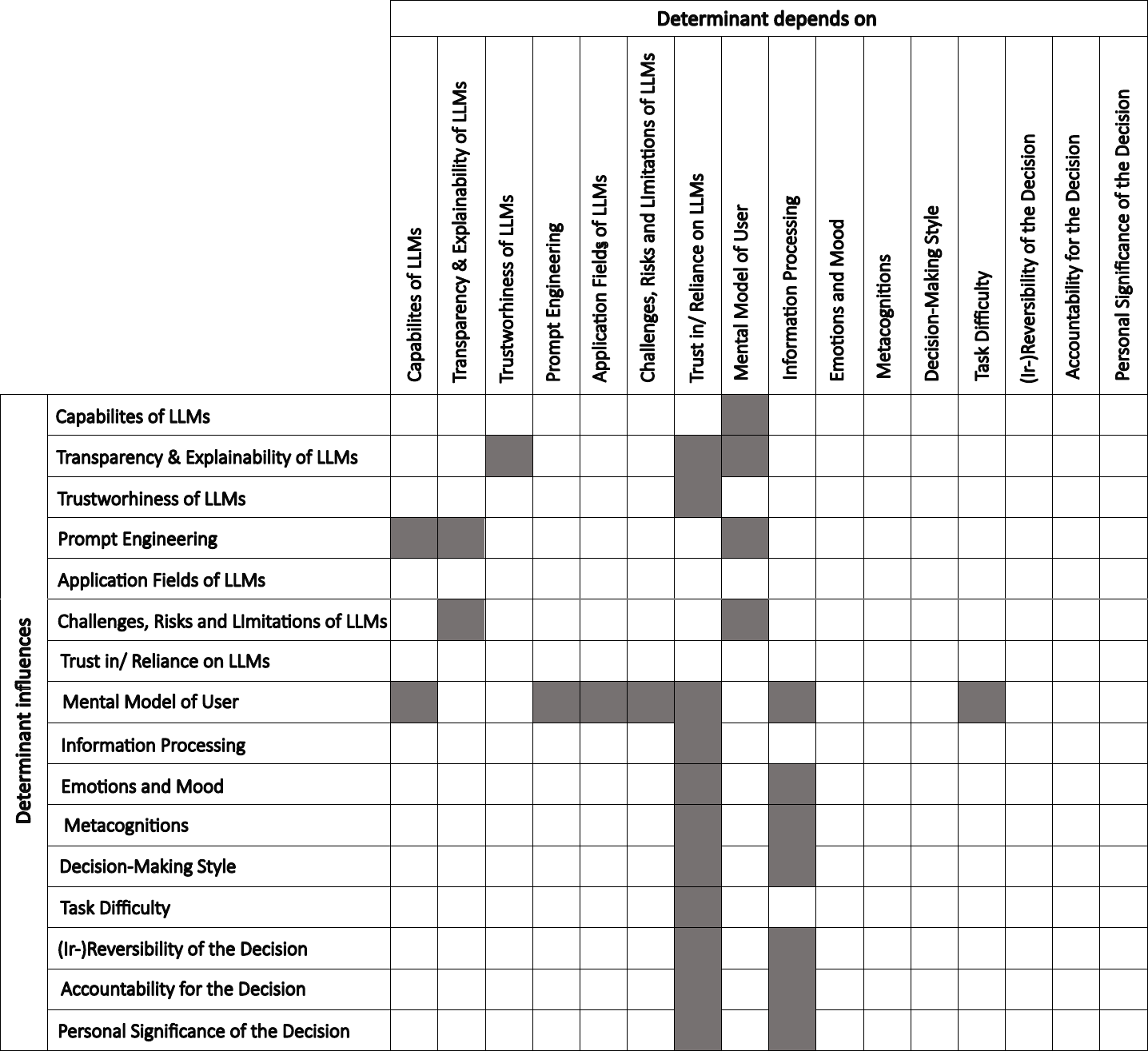}
	\caption{Matrix of reciprocal dependencies between determinants of LLM-assisted decision-making.}
	\label{fig:Matrix Determinants LLM-supported Decision Making}
\end{figure*}

Drawing from our literature review spanning various disciplines (such as psychology and artificial intelligence) as well as our comprehensive analysis, we introduce a \textit{dependency framework} of determinants of LLM-assisted decision-making. This framework systematizes the potential interactions between technological, psychological, and decision-specific factors in terms of reciprocal interdependencies. The resulting dependency framework is illustrated as a feature diagram (see Fig.\ \ref{fig:Determinants LLM-supported Decision Making}; for notation details, also refer to Section \ref{subsec: Determinants Structure}) and as a dependency matrix (see Fig.\ \ref{fig:Matrix Determinants LLM-supported Decision Making}).
In particular, Figure \ref{fig:Determinants LLM-supported Decision Making} gives a structural overview of all determinants considered in this paper, amended by identified interactions between the determinants in terms of interdependencies. The details on these dependencies are provided within the determinant-specific analyses in Sections \ref{sec:tech-determinants}, \ref{sec:psych-determinants}, and \ref{sec:ds-determinants}. 

In a comparison of technological, psychological, and decision-specific determinants, it is noteworthy that the greatest number of interactions is observed among the psychological factors. In contrast to technological and psychological factors, there are the least interactions with other factors at the decision-specific level. The dependencies illustrated in Figure \ref{fig:Determinants LLM-supported Decision Making} demonstrate that trust in or reliance on LLMs in decision-making processes is influenced by the greatest number of factors. For example, emotions and mood impact trust in LLM-assisted decision-making, as discussed in Section \ref{subsec: Emotions and Mood}. Similarly, the user's mental model, representing the cognitive understanding of how LLMs function and relevant experiences, also plays a crucial role in influencing trust in this technology as a support in decision-making (see Section \ref{subsec: Mental Model}). Another factor affecting reliance on LLMs in decision-making is the type of information processing, which can be either slow and deliberate or fast and intuitive (see Section \ref{subsec: Emotions and Mood}).  Not only psychological factors but also technological determinants, such as the transparency or the trustworthiness of LLMs can exert an impact on trust in or reliance on LLMs within human decision-making processes. Moreover Figure \ref{fig:Determinants LLM-supported Decision Making} indicates that, according to the results of the literature analysis, trust or reliance might not impact the other determinants examined in this study.

As evident from Figure \ref{fig:Determinants LLM-supported Decision Making}, the mental model of the user interacts with a variety of other factors in the context of LLM-assisted decision-making (see Section \ref{subsec: Mental Model}). Unlike trust in and reliance on LLMs, the mental model is not only impacted by other factors but also exerts influence on other determinants. For instance, psychological factors like information processing affect the user's mental model through cognitive biases. Additionally, on a technological level, the mental model of the user is also impacted, for example, by integrating prompting techniques into its cognitive representation. On the other hand, the user's mental model influences prompting, as it encompasses knowledge and experiences related to prompting techniques. At the decision-specific level, the user's mental model, representing their expertise, influences the perceived difficulty of the decision situation.

Furthermore, as elaborated in Section \ref{subsec: Information Processing}, information processing plays a crucial role in LLM-assisted decision-making, as indicated by the number of interactions in Figure \ref{fig:Determinants LLM-supported Decision Making}. Among the psychological factors, information processing is, for instance, influenced by emotions or mood. Concerning decision-specific determinants (see Section \ref{sec:ds-determinants}), the nature of information processing is determined by factors such as the personal significance and the (ir-) reversibility or the decision. Information processing, in turn, influences the extent to which individuals exhibit reliance on LLMs.

In Figure \ref{fig:Matrix Determinants LLM-supported Decision Making}, the factors influencing LLM-assisted decision-making are presented in a two-dimensional dependency matrix. For each determinant considered in the paper, the matrix illustrates the reciprocal influence between that determinant and others, showcasing which factors it influences and in turn which determinants influence it.  By analyzing the matrix, it becomes apparent, for instance, that prompt engineering can exert an impact on the capabilities of LLMs, their transparency and explainability, and the user's mental model within the realm of LLM-assisted decision-making. Conversely, prompt engineering is subject to influence by the user's mental model, encompassing their knowledge and experiences.

\section{Discussion}
\label{sec:discussion}
In this section, we first discuss the implications of the analyzed determinants with regard to the efficacy of LLM-assisted decision-making for both individuals and organizations (Section \ref{subsec:implications}). Furthermore, we address criticisms and limitations inherent to the present work (Section \ref{subsec:limitations}), and finally propose avenues for future research (Section \ref{subsec:futureWork}).

\subsection{Implications for Decision-Makers and Organizations}
\label{subsec:implications}
In the field of decision-making, LLMs have emerged as valuable assets for both organizations and individuals (refer to Section \ref{subsec: LLM-assisted DMP}). Nonetheless, incorporating LLMs into decision-making processes comes with its own set of complexities and challenges (see Section \ref{subsec: Challenges}). Hence, it is crucial to understand the technological (see Section \ref{sec:tech-determinants}), psychological (see Section \ref{sec:psych-determinants}), and decision-specific factors (see Section \ref{sec:ds-determinants}), along with their interactions (see Section \ref{sec:framework}) to effectively analyze, evaluate, and enhance the effectiveness of decision-making processes. This understanding facilitates the explanation and prediction of user behavior in the context of LLM-assisted decisions. For example, the quality of decisions assisted by LLMs can be affected by either over-reliance or under-reliance. This, in turn, may be influenced by various factors, such as information processing or emotional states.

To enhance the efficiency and effectiveness of LLM-assisted decision-making within organizational contexts it is essential to implement comprehensive training programs. These programs should aim to improve users' comprehension and interaction with LLMs. The training should encompass detailed insights into the functional capabilities (refer to Section \ref{subsec: Capabilites}) and inherent constraints of LLMs (see Section \ref{subsec: Challenges}). Such educational programs should focus on clarifying the processes by which these models analyze and generate information. Another critical component of this training involves the development of analytical skills that enable individuals to identify and critically evaluate potential biases, inaccuracies, or gaps in the information produced by LLM-generated outputs. Additionally, a crucial aspect of the training is the effective formulation of prompts, as this directly impacts the relevance and accuracy of the model's responses. This ensures that the responses of the LLMs are aligned with the specific context and requirements of the decision under consideration.

In the realm of LLM-assisted decision-making, mitigating the risk of over-reliance (refer to Section \ref{subsec:trustReliance}) requires a comprehensive approach, encompassing measures at both individual and organizational levels. At the individual level a key factor involves developing an awareness of potential biases (see Section \ref{subsec: Information Processing}) that can lead to over-reliance in decision-making. Training should not only identify these common biases but also provide practical techniques for recognizing and addressing them in real-world decision-making situations. Consequently, the training should include strategies designed to disrupt automatic, heuristic-based decision-making processes, thereby encouraging individuals to engage in more thoughtful, reflective, and analytical thinking. At the organizational level, fostering a culture that values and encourages critical examination of decisions assisted by LLMs is of utmost importance. Additionally, the regular monitoring and evaluation of LLM-assisted decisions are crucial. These practices are key in helping organizations identify and address trends of over-reliance and also under-reliance on LLMs. Furthermore, the integration of feedback mechanisms within the organizational structure is essential. These mechanisms should enable employees to report any discrepancies or biases they observe in LLM-generated outputs. The feedback collected is vital for the iterative refinement of LLMs, significantly enhancing their accuracy and reliability. Moreover, this information can be utilized to establish a centralized feedback platform or repository. This repository can act as a comprehensive database, ensuring that insights and concerns raised by employees are transparently and efficiently communicated across various organizational levels.The maintenance of this repository requires regular updates, focusing on two main aspects. Firstly, it should systematically document the feedback provided by employees, capturing their observations, experiences, and critiques related to the LLM-generated outputs. Secondly, the repository should also maintain detailed records of the actions taken in response to the feedback. 

Moreover, under-reliance, as discussed in Section \ref{subsec:trustReliance}, on LLMs in decision-making processes presents a critical challenge. Reducing under-reliance on LLMs requires a comprehensive strategy that addresses this issue at both the individual and organizational levels. For instance, at an individual level, training programs could strive for a better understanding of LLM's capabilities (refer to Section \ref{subsec: Capabilites}) and limitations (see Section \ref{subsec:limitations}), fostering awareness about scenarios where under-reliance may compromise decision quality, and identifying optimal situations for LLM integration in decision-processes. At the organizational level, cultivating a culture that embraces LLMs as collaborative decision support tools is paramount. Furthermore, the formulation of policies or guidelines regarding the consideration of LLM-generated recommendations might be supportive in mitigating the risk of under- and over-reliance.

Integrating emotional awareness training is pivotal for enhancing decision quality in the domain of LLM-assisted decision-making. The training should emphasize understanding the potential impact of various emotional states on decisions made with the assistance of LLMs (refer to Section \ref{subsec: Emotions and Mood}). The program should begin with an introduction to theories that link emotions to decision-making processes. This theoretical foundation aids participants in understanding how emotional states can influence their interactions with LLMs and influence the decision-making process. Another significant aspect of the training involves teaching individuals to recognize their own emotional states and understand how these states may bias their interpretation of LLM-generated outputs and the decision-making process. Additionally, the training should guide participants on strategies to manage the influence of emotions on their decision-making. Incorporating practical exercises and scenario-based learning, where participants can apply their understanding in simulated decision-making situations involving LLMs, is essential. An integral part of the training should be dedicated to feedback and reflection sessions. In these sessions, participants could discuss their experiences, challenges, and insights gained during the exercises.

Similarly, as discussed in Section \ref{subsec: Metacognitions}, understanding and managing metacognitions, such as the Feeling of Rightness (FOR) and the Feeling of Error (FOE), is significant to improve LLM-assisted decision-making. For example, by recognizing the signs of over-reliance associated with a strong FOR employees can learn to critically analyze their reliance on LLM outputs with their judgment. Such a training is vital in preventing over-reliance on LLMs, especially in situations where a strong FOR might lead to uncritical acceptance of LLM suggestions.

In the domain of LLM-assisted decision-making, the implementation of customized training programs tailored to distinct decision-making styles emerges as an important strategy for enhancing decision quality (see Section \ref{subsec: Decision-making styles}). Recognizing the diverse approaches individuals take in decision-making – categorized as minimizers, satisficers, and maximizers – these training modules should aim to address the unique challenges and tendencies associated with each style. For minimizers, who typically prioritize efficiency and speed, training programs should emphasize the importance of not solely relying on the first satisfactory option presented by LLMs. These programs should encourage a more comprehensive evaluation of LLM-generated outputs, highlighting the potential risks of over-reliance on initial suggestions. Satisficers, known for seeking options that meet specific criteria, require training that focuses on effectively leveraging LLMs to find options that align with their criteria. It should guide satisficers in using LLMs as a tool for narrowing down choices while still engaging in a critical evaluation of the options presented. For maximizers, who endeavor to achieve the best possible outcome through extensive research and evaluation, training should equip them with strategies to efficiently harness the vast information processing capabilities of LLMs while maintaining a critical perspective.

Both individuals and organizations are advised to approach LLM-assisted decision-making with a mindset that values critical analysis, especially in situations involving irreversible decisions or those of high personal or organizational significance. In the sphere of organizational decision-making, particularly when decisions are perceived as irreversible or bear substantial consequences, it is necessary for organizations to establish and enforce policies or guidelines that actively promote a culture of critical evaluation of suggestions derived by LLMs, especially in scenarios where the stakes are high.

\subsection{Limitations}
\label{subsec:limitations}
This section addresses limitations of the present work exploring determinants of LLM-assisted decision-making. 

Literature reviews are often susceptible to bias. For instance, the selection process in a literature review can introduce biases. Selection bias occurs when the articles chosen for a review do not represent the entire evidence base \cite{McDonagh2013AvoidingBI}. Publication bias means that research results that are statistically significant are more frequently published compared to those not statistically significant \cite{Mlinari2017DealingWT}. Consequently, literature reviews relying solely on traditional, commercially published academic research will likely exhibit the same level of bias as the research they are based on \cite{Haddaway2020EightPW}. Given the limited number of studies to date related to LLM- or AI-assisted decision-making identified by our review, the potential for selection bias and publication bias in this research domain cannot be denied.

A further limitation is that our literature review did not systematically conduct the collection and synthesis of previous research \cite{Snyder2019Literature}. It should be noted, however, that the primary objective of the present work was to integrate the current state of knowledge through an integrative literature review to generate new knowledge \cite{holton2002mandate}; also see Section \ref{sec:approach-overview}. In the context of an integrative literature review, the search strategy is usually not performed in a systematic way \cite{Snyder2019Literature}. 

The rapid development of LLMs \cite{Chang2023ASO} poses a considerable challenge to the relevance and accuracy of research in this field. Thus, studies conducted even a few months prior may not accurately reflect the current state of technology and understanding of LLMs. Consequently, conclusions drawn from these studies risk becoming outdated shortly after their publication. However, we argue that despite the advancements in LLM technology, the fundamental psychological mechanisms, such as information processing, the significance of emotional states, and mental models, that govern how individuals make decisions with the assistance of LLMs, are likely to remain stable. Similarly, the influence of decision-specific determinants, such as their importance to the decision-maker or their reversibility, on LLM-assisted decision-making is expected to remain relatively consistent.

The exclusion of organizational determinants of LLM-assisted decision-making represents a further limitation in the current work. Organizational factors are recognized for their impact on individual decision-making processes \cite{Barbera2019Theweight}. For instance, one factor related to the organizational context involves “articulated and often informal rules-of-thumb shared by multiple participants within the firm” \cite[p. 31]{Bingham2007What}. These rule-of-thumbs might influence individuals' decisions assisted by LLMs. Employees' perception and trust in LLMs may be impacted by these articulated and informal rules. If the prevailing sentiment within the organization tends towards skepticism of AI and new technologies, it could result in a general reluctance to rely on LLMs for decision-making. On the other hand, an organizational culture that embraces technological solutions might foster a more trusting attitude towards LLMs, encouraging their use in various decision-making processes. Nonetheless, we have excluded organizational factors because they are beyond the direct control of individuals. Additionally, in our literature review, we have not considered environmental factors, as they are, unlike technological or psychological factors, relatively difficult for decision-makers or organizations to influence.

Another limitation is that the current literature review does not encompass the entirety of technological, psychological, and decision-specific determinants involved in LLM-assisted decision-making. For example, in the area of decision-specific determinants, time pressure might play a significant role in LLM-assisted decision-making. The amount of time available for making a decision might impact the reliance on LLMs, with increased pressure potentially leading to faster, less deliberative decision-making. 

In this paper, we have conducted a comprehensive literature review to identify potential influences of psychological, technological, and decision-specific factors on decision-making processes supported by LLMs. Our analysis has focused on extracting and synthesizing information from existing literature to understand how these diverse factors might interact and impact LLM-assisted decision-making. However, it is important to note that the majority of these influences and interactions identified in our review are predominantly theoretical and based on hypothesized interactions, rather than being grounded in empirical data. This reliance on theoretical frameworks and hypotheses, rather than empirical evidence, presents a limitation in our study. The absence of direct empirical validation means that our conclusions about the interaction and impact of these factors on LLM-assisted decision-making should be interpreted with caution.

A further constraint of our study is that it exclusively focuses on decision-making supported by traditional unimodal LLMs without considering Multimodal Large Language Models (MLLMs). While unimodal LLMs can only process natural language \cite{Wu2023Multimodal}, MLLMs have the ability to process multimodal information, such as images or videos. Hence, MLLMs are more in line with the way individuals perceive the world multisensorily. Consequently, MLLMs could enable users to interact more flexibly with intelligent assistants based on multimodal inputs and support them in a broader spectrum of tasks \cite{yin2023survey}. However, our work does not account for the potential differences in decision-making processes facilitated by MLLMs compared to unimodal LLMs.

\subsection{Future Work}
\label{subsec:futureWork}
Drawing from the implications and limitations identified in Section \ref{subsec:limitations}, several future research directions can be outlined.

A significant direction for future research in the domain of LLM-assisted decision-making involves conducting a systematic literature review on the influencing factors of LLM-assisted decision-making. This entails identifying, critically evaluating, and collecting data from relevant research \cite{Liberati2009e1} and synthesizing research findings in a systematic, transparent, and reproducible manner \cite{Davis2014Viewing}. The application of a systematic literature review aids in reducing biases, for instance, by identifying empirical evidence that aligns with predefined inclusion criteria, thereby generating reliable findings from which meaningful conclusions can be drawn \cite{Moher2010Preferred}.

A critical area of future research involves empirically examining the postulated effects of and interactions among psychological, technological, and decision-specific determinants. To determine whether the identified determinants of LLM-assisted decision-making constitute causal factors, it is essential for future studies to employ experimental designs to examine whether variations in these factors systematically affect the decision-making behavior of users. For instance, potential research could investigate how different emotions impact the acceptance of recommendations generated by LLMs in decision-making, as well as the extent of reliance on LLM-generated suggestions, by utilizing experimental design methodologies. To evoke the desired emotions, film footage, for example, can be utilized \cite{Hu2014TheEffect}, which has been proven to be an effective method to induce emotions \cite{Forgas19f87After, Liu2010TheEffects}.

As described in Section \ref{sec:framework}, in LLM-assisted decision-making, trust in or reliance on LLMs is likely to be a determinant influenced by a variety of other factors. Moreover, the relatively high number of interactions involving the user's mental model and information processing with other determinants emphasizes their significance in decision-making processes facilitated by LLMs and their inherent complexity. Future research should address these determinants of LLM-assisted decision-making and their interactions with other factors to comprehensively understand their impact. 

Our literature review reveals a lack of consideration for organizational determinants in the realm of LLM-assisted decision-making. Consequently, future research should focus on exploring the dynamics between technological, psychological, decision-specific, and organizational determinants to provide a more comprehensive understanding of their impact on LLM-assisted decision-making. For instance, organizational norms regarding the use of intelligent systems can influence the perceived usefulness of such systems, subsequently affecting the intention to use artificial intelligence \cite{VormIntegrating2022}. Therefore, organizational norms, potentially governing acceptable and expected LLM usage within an organization, may significantly influence how LLMs are incorporated into the decision-making process. Understanding and analyzing these norms can offer valuable insights into the facilitators and barriers to LLM adoption in organizational contexts.

Moreover, future research could concentrate on the development and evaluation of training programs, which are essential for optimizing LLM-assisted decision-making. For example, such programs can provide users with a comprehensive understanding of LLM capabilities and limitations, thereby establishing realistic expectations for their applications. Additionally, the training could emphasize awareness of psychological factors that influence decision-making with LLMs, as well as critical analytical skills, ensuring that decisions are thoroughly informed and not merely based on LLM-generated outputs. It is crucial to evaluate these programs to determine their effectiveness and to make necessary adjustments.

A compilation of real-world cases, akin to the collection curated by Urbach and Roeglinger \cite{Urbach2019Introduction}, focusing on the application of LLM-assisted decision-making across diverse organizations and domains, can serve as a proposal for future research endeavors. This compilation could provide valuable insights, best practices, and lessons learned from various organizations, illustrating how they have proficiently overcome challenges and capitalized on opportunities in the decision-making process supported by LLMs. As a subsequent step, a systematic analysis of these cases could be conducted to identify methods and techniques that enhance the effective utilization of LLMs in human decision-making. Specifically, these cases could be systematically categorized based on factors such as decision complexity, the domain of application (e.g., healthcare, finance, legal), the nature of the LLM’s involvement (e.g., data analysis, prediction, recommendation), the psychological factors involved and the outcomes achieved. Such a systematization would facilitate a more profound understanding of the contexts in which LLMs are most effective, the types of decisions that benefit most from LLM assistance, and the potential pitfalls or challenges encountered.

Our paper focuses on traditional unimodal Large Language Models (LLMs), which are trained and applied to text data \cite{Wu2023Multimodal}. In contrast, Multimodal Large Language Models (MLLMs) possess the capability to process multimodal information, such as images and audio, aligning more closely with human perception and cognition, potentially enhancing human decision-making processes \cite{yin2023survey}. Therefore, investigating determinants of decision-making processes supported by MLLMs presents an important area for further research. The aim of future studies could be to explore whether similar, additional, or distinct determinants underlie the decision-making process supported by multimodal models (MLLMs) compared to unimodal models (LLMs).

Subsequent research initiatives can focus on multi-agent collaborations to explore influencing factors in the context of LLM-assisted decision-making. In contrast to single LLM agents, multi-agent systems provide enhanced capabilities through collaborations among diverse LLM-powered agents, each equipped with specialized abilities and a distinct role \cite{han2024llm, jinxin2023cgmi, zhang2023exploring}. These systems hold significant potential to simulate complex real-world scenarios, facilitating interactions among the diverse agents engaged in planning, discussions and decision-making, reflecting the cooperative dynamics of problem-solving tasks in human group work \cite{guo2024large}. For example, the determinants of the triadic interplay among the principal decision-making entities, specifically human users, LLM-powered agents, and the governing mechanisms or rules embedded in the systems \cite{handler2023balancing}, can be systematically investigated to enhance LLM-assisted decision-making by understanding and optimizing the interactions within these entities during the decision-making process.

\section{Conclusion}
\label{sec:conclusion}
In conclusion, this work presents a comprehensive analysis of determinants impacting human decision-making assisted by Large Language Models (LLMs). As described in Section \ref{sec:tech-determinants}, we explored technological aspects of LLMs, including transparency and prompt engineering, and their influence on decision-making processes. Furthermore, we examined psychological determinants, such as the impact of emotions, metacognitions, and decision-making styles on the efficacy of LLM-assisted decisions, for details, refer to Section \ref{sec:psych-determinants}. Additionally, we addressed decision-specific factors such as task complexity and accountability for the decision, as discussed in Section \ref{sec:ds-determinants}. 
For each determinant, its impact on LLM-assisted decision making and potential interactions with other determinants are analyzed in detail. This analysis is accompanied by a scenario-based illustration of the determinant's potential influence on the decision-making process. 

Building upon our literature review and analysis, our study not only provides a structural overview of technological, psychological and decision-specific factors influencing decision-making assisted by LLMs, but also a \textit{dependency framework} which systematizes and visualizes the reciprocal interdependencies among the various determinants, as detailed in Section \ref{sec:framework}. For instance, our work indicates that, owing to their diverse interactions with other determinants, factors such as trust in or reliance on LLMs (see Section \ref{subsec:trustReliance}), the user's mental model (refer to Section \ref{subsec: Mental Model}), along with the nature of information processing (see Section \ref{subsec: Information Processing}), are likely key aspects in LLM-assisted decision-making.

The resulting structural overview and dependency framework of determinants can facilitate the explanation and prediction of user behavior in the context of LLM-assisted decision-making. This can, for instance, augment the awareness of users and organizations concerning potential risks inherent in LLM-assisted decisions, such as those arising from limitations in LLMs (see Section \ref{subsec:limitations}) or over-reliance (refer to Section \ref{subsec:trustReliance}). Consequently, appropriate measures can be implemented to mitigate these risks, thereby enhancing decision quality. Furthermore, understanding the influence of factors on human decision-making behavior with LLM support empowers users and organizations to optimize and capitalize on the advantages of LLM-assisted decision-making. For example, as outlined in Section \ref{subsec: Prompt Engineering}, proficiency in prompt engineering can improve the transparency of LLMs, enabling users to assess the LLM-generated output more critically and potentially leading to improved decision-making. Moreover, as discussed in Section \ref{subsec:futureWork}, this work establishes a foundation for future empirical studies, including experimental investigations on factors influencing LLM-assisted decision-making, such as the impact of emotions and mood.

\section*{Acknowledgements}
The authors gratefully acknowledge the support from the "Gesellschaft für Forschungsförderung (GFF)", as this research was conducted at Ferdinand Porsche Mobile University of Applied Sciences (FERNFH) as part of the "Digital Transformation Hub" project funded by the GFF with means of the Province of Lower Austria.

\bibliographystyle{abbrv}
\bibliography{references}

\end{document}